%% file: buniming.tex
\documentclass[11pt]{article}

\usepackage[utf8]{inputenc}
\usepackage[T1]{fontenc}
\usepackage{times}

\usepackage[margin=1in]{geometry}

\usepackage{amsmath,amssymb,amsfonts,bbm}
\usepackage{algorithm}
\usepackage{algpseudocode}
\usepackage{graphicx}
\usepackage{booktabs}
\usepackage{multirow}
\usepackage{longtable}
\usepackage{array}
\usepackage[table]{xcolor}
\usepackage{wrapfig}
\usepackage{caption}
\usepackage{float}
\usepackage{placeins}
\usepackage{natbib}
\usepackage{hyperref}
\usepackage{url}

\definecolor{hmecobg}{RGB}{235,244,252}

\definecolor{promptvariable}{RGB}{155,45,45}

\newsavebox{\promptbox}
\newenvironment{prompttemplate}[1]{%
  \par\medskip\noindent\begin{minipage}{\linewidth}
  \centering\normalsize #1\par\smallskip
  \setlength{\fboxsep}{8pt}
  \setlength{\fboxrule}{0.4pt}%
  \begin{lrbox}{\promptbox}
  \begin{minipage}{\dimexpr\linewidth-2\fboxsep-2\fboxrule\relax}
  \fontsize{8}{10}\selectfont\ttfamily\raggedright
  \setlength{\parindent}{0pt}
  \setlength{\parskip}{2pt}%
}{%
  \end{minipage}\end{lrbox}%
  \fcolorbox{black!65}{black!3}{\usebox{\promptbox}}%
  \end{minipage}\par\medskip
}

\date{October 2026}

\title{Reflect After Failure: Brain-Inspired Self-Evolving Agents for Embodied Spatial Intelligence}

\author{
Xin Li\textsuperscript{1,*} \quad
Xiaohuan Gu\textsuperscript{2,*} \quad
Yifan Wang\textsuperscript{3} \quad
Shaobin Ling\textsuperscript{4} \quad
Tianli Shi\textsuperscript{1} \quad
Jingao Xu\textsuperscript{5} \\
Zhimeng Yin\textsuperscript{1} \quad
Dapeng Wu\textsuperscript{1} \quad
Yun Wang\textsuperscript{1,\textdagger} \\
\textsuperscript{1}City University of Hong Kong \quad
\textsuperscript{2}East China Normal University \\
\textsuperscript{3}University of Electronic Science and Technology of China \\
\textsuperscript{4}The Chinese University of Hong Kong, Shenzhen \quad
\textsuperscript{5}The University of Hong Kong \\
\textsuperscript{*}Equal contribution. \quad
\textsuperscript{\textdagger}Corresponding author.
}

\begin{document}
\maketitle

\begin{abstract}
Vision-Language Models (VLMs) have made substantial progress in spatial reasoning, and self-evolving agents further extend these capabilities to embodied environments by learning from interactions. However, current self-evolution remains particularly challenging in embodied spatial tasks, complex and multi-step decisions from partial visual evidence making it difficult to determine why an execution fails. The failure may stem from misperception, insufficient exploration, or incorrect planning, yet execution feedback alone rarely reveals which component is responsible. As a result, simply accumulating interaction experience can reinforce ineffective strategies. To address this, we propose HiMeco, a brain-inspired self-evolving agent for embodied spatial intelligence, inspired by metacognitive monitoring and hippocampal memory consolidation. Through Hippocampal-Metacognitive Reflection, HiMeco diagnoses failures in perception-exploration-planning (PEP), then compares execution attempts to derive reusable strategies and guidance for avoiding recurring errors. It also abstracts interaction programs into parameterized skills, using evidence contracts and execution-based validation to determine their retention and reuse. These strategies and skills guide subsequent interactions, with memory updated only between rounds and model parameters kept frozen. HiMeco achieves state-of-the-art performance on ESI-Bench, and it also improves accuracy on EmbSpatial-Bench by 3.94 percentage points without evolution.
\end{abstract}

\section{Introduction}

Embodied agents operating in complex physical environments require robust spatial intelligence, the ability to not only perceive visible geometry but also reason about hidden structures and topologies \citep{majumdar2024openeqa,yang2025thinking,gong2026space10,feng2026seeing}. While Vision-Language Models (VLMs) have demonstrated remarkable zero-shot reasoning capabilities, their spatial understanding remains fundamentally constrained by passive observation \citep{yang2025thinking,gong2026space10,hong2026esibench}. In real-world scenarios riddled with non-line-of-sight (NLOS) occlusions and unobservable physical dynamics, passive VLMs struggle to perform deep spatial reasoning without actively interacting with the environment to uncover latent topologies \citep{feng2026seeing,majumdar2024openeqa,hong2026esibench}.

Existing research on enhancing spatial reasoning in VLMs broadly falls into two paradigms based on physical feedback availability: static perception \citep{cheng2024spatialrgpt,song2025robospatial,zhu2025llava3d} and dynamic interaction \citep{ziliotto2025tango,hong2026esibench}. The former, whether through offline optimization (e.g., SFT or RL) or static prior injection from expert modules (e.g., 3D/depth estimation and prompt engineering), treats the environment as a passive and fixed observation space. Lacking closed-loop interaction, it relies heavily on explicit in-view visual cues, suffering from cognitive blind spots under NLOS occlusions and hidden topologies \citep{feng2026seeing,hong2026esibench}. In contrast, the dynamic interaction paradigm incorporates physical environment feedback into planning, enabling agents to actively resolve local ambiguities \citep{ziliotto2025tango,kim2025flare}. However, extending self-evolving agent to embodied spatial tasks introduces a critical bottleneck. Prior work on self-evolving agent often focused on tasks where the causes of failure are relatively easy to identify. While in embodied spatial tasks, agents must make complex multi-step decisions based on partial and dynamically acquired visual evidence in continuous 3D environments subject to physical noise. These conditions make failures difficult to diagnose, while feedback across interaction rounds offers limited guidance on how to revise the underlying strategies \citep{li2024eai,duan2025aha,qi2026armor,ju2026embodiskill}. This ambiguity hinders agents from reliably assessing which strategies are effective and identifying productive directions for further adaptation. As a result, agents may reinforce flawed experiences and update strategies in the wrong direction, causing errors to accumulate over successive evolution rounds \citep{ju2026embodiskill}.

Remarkably, animals can efficiently distill effective strategies from experience to solve embodied spatial tasks in unfamiliar environments. This adaptive ability of biological brains is supported in part by interactions between the hippocampus and prefrontal metacognitive circuits \citep{liao2024learning,jensen2024recurrent,fleming2026integrative}. In contrast, existing embodied spatial intelligence, whether static (e.g., via parameter updates or external tools) or dynamic self-evolving systems, struggle to deliver accurate spatial reasoning during active exploration \citep{yang2025thinking,feng2026seeing,hong2026esibench}. Because they cannot determine whether failures stem from noisy perception, suboptimal exploration, or improper planning \citep{li2024eai,ju2026embodiskill}. This makes it difficult to translate interaction feedback and accumulated experience into consistent behavioral improvements, leaving errors unresolved across successive interactions \citep{ju2026embodiskill,zhang2026sma,tian2026nesy}. These limitations motivate Hippocampal-Metacognitive Reflection, which couples failure diagnosis with memory consolidation to guide future interactions. Inspired by these neurobiological functions, we hope to achieve a brain-inspired self-evolving agent equipped with multi-turn Hippocampal-Metacognitive reflection for resource-efficient and adaptive embodied spatial intelligence.

\begin{figure*}[t]
\centering
\includegraphics[width=\textwidth]{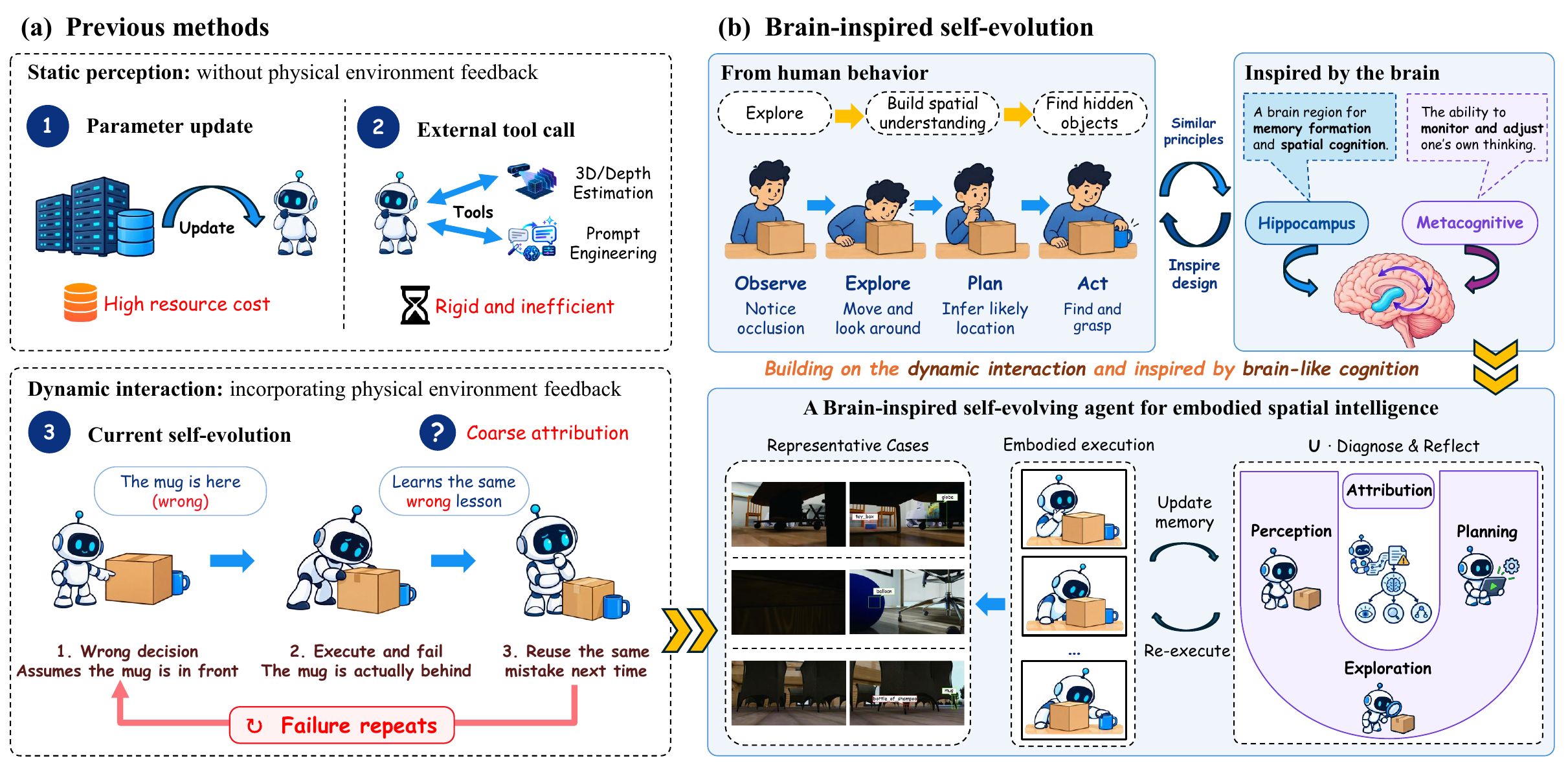}
\caption{\small
\textbf{\textit{Motivation of our method.}}
\textbf{\textit{(a)}} Previous methods risk reinforcing errors through inaccurate failure attribution.
\textbf{\textit{(b)}} Inspired by metacognitive monitoring and hippocampal consolidation, our method diagnoses failures to guide future interactions.
}
\label{fig:motivation}
\end{figure*}

Translating this brain-inspired concept into practice presents two challenges. The first is adaptive semantic abstraction: deciding what to retain from repeated interactions and how broadly it should apply \citep{zhang2026sma,shu2026remem,xu2026multigranularity}. Across interaction and evolution rounds, a parameter-frozen agent must turn individual attempts into reusable strategies and executable skills. Overly specific knowledge transfers poorly, while overly general knowledge loses the detail needed for reliable execution \citep{wang2025awm,zhang2026sma}. The second is reliable failure attribution \citep{li2024eai,ju2026embodiskill}. An incorrect answer alone does not reveal whether the agent misinterpreted its observations, explored insufficiently, or executed a flawed plan. Without this distinction, the agent may retain unreliable procedures or reinforce ineffective strategies.

To address these challenges, we propose \textsc{HiMeco}, a brain-inspired self-evolving agent for embodied spatial reasoning. It consolidates interaction experience into a strategy bank $\mathcal{E}$ and a library of parameterized executable skills $\mathcal{K}$. Evidence contracts and invocation-based validation govern which skills are retained, promoted, or suspended across evolution rounds. A perception-exploration-planning (PEP) diagnostic module $\mathcal{U}$ uses execution traces, task outcomes, and privileged simulator metadata to assess failures across these three dimensions after execution. These diagnostics guide candidate comparison and reflection, yielding reusable strategies and negative constraints for subsequent interactions. Model parameters remain frozen, and both memories are updated only between rounds.

\textsc{HiMeco} alternates between online inference and offline evolution. During inference, the frozen VLM generates programs conditioned on retrieved experiences and skills, while evidence contracts check the observations and intermediate outputs used for reasoning. Memory remains fixed throughout each interaction round. At the round boundary, diagnostic feedback guides comparative reflection and experience consolidation, while execution outcomes support skill abstraction and lifecycle updates. The resulting memories guide subsequent interactions without modifying model parameters. This separation allows the agent to accumulate experience across rounds while evaluating each attempt against a fixed memory snapshot.

Our contributions are summarized as follows:

\begin{itemize}
\item We propose \textsc{HiMeco}, a brain-inspired self-evolving agent for embodied spatial reasoning, which repeats interaction and memory evolution without updating model parameters.
\item We develop a Hippocampal-Metacognitive Reflection mechanism that couples PEP diagnostic module with the consolidation of reusable strategies and parameterized skills.
\item We comprehensively evaluate on ESI-Bench for active embodied spatial reasoning, \textsc{HiMeco} demonstrating state-of-the-art performance. It also improves accuracy on EmbSpatial-Bench without evolution, showing its effectiveness in passive settings.
\end{itemize}

\section{Related Work}

\paragraph{Embodied Spatial Intelligence.}
Benchmarks span static visual reasoning, such as EmbSpatial-Bench \citep{du2024embspatial}, and interactive evaluation, such as ESI-Bench \citep{hong2026esibench}, which requires acquiring evidence through actions. Existing approaches use task-specific fine-tuning \citep{song2025robospatial,wang2025metavqa}, explicit 3D memory \citep{zeng2026janusvln,zheng2025esceme}, or interactive exploration \citep{ziliotto2025tango,hong2026esibench} to improve spatial understanding. Training-free alternatives organize frozen models within external structure: TANGO composes navigation and perception primitives into task-specific programs without additional training \citep{ziliotto2025tango}, NeSy-Spatial abstracts tool interactions into typed and composable spatial skills that self-evolve with use \citep{tian2026nesy}, and Spatial Memory Agent consolidates verified spatial experience into transferable procedural lessons \citep{zhang2026sma}. However, effective interaction alone does not ensure that agents can diagnose failures and consolidate experience into reusable strategies and skills \citep{hong2026esibench,ju2026embodiskill,zhang2026sma}.

\paragraph{Experience-to-Skill Abstraction.}
Self-evolving agents adapt by storing and reusing interaction experience, often without updating model parameters \citep{wang2025awm,liu2025cer,xu2025amem,zhang2026memgen}. VOYAGER accumulates an ever-growing library of executable code skills under an automatic curriculum \citep{wang2023voyager}, and later work evolves increasingly broader portions of the agent system: SHAPER co-evolves reusable skills and a context-code harness around a frozen planner \citep{wang2026shaper}, MetaSkill-Evolve makes skill improvement itself recursive by evolving meta-skills on a slower timescale \citep{wang2026metaskill}, and the Darwin Gödel Machine applies open-ended evolution to an archive of self-modifying agents \citep{zhang2026dgm}. Complementarily, Training-Free GRPO distills group-relative feedback into experiential token priors rather than parameter updates \citep{youtu2025tfgrpo}. These frameworks refine textual guidance or execution programs through reflection, with applications in code generation and web navigation.In embodied environments, this process is complicated by perceptual errors,insufficient exploration, and planning failures, and skill-aware reflectionhas begun to separate flawed skill content from execution lapses\citep{ju2026embodiskill}.Without reliable attribution and execution-based validation, agents risk
retaining ineffective strategies or unreliable skills.Existing skill-evolution frameworks rarely distinguish these causes before
updating memory.

\begin{figure*}[t]
\centering
\includegraphics[width=\textwidth]{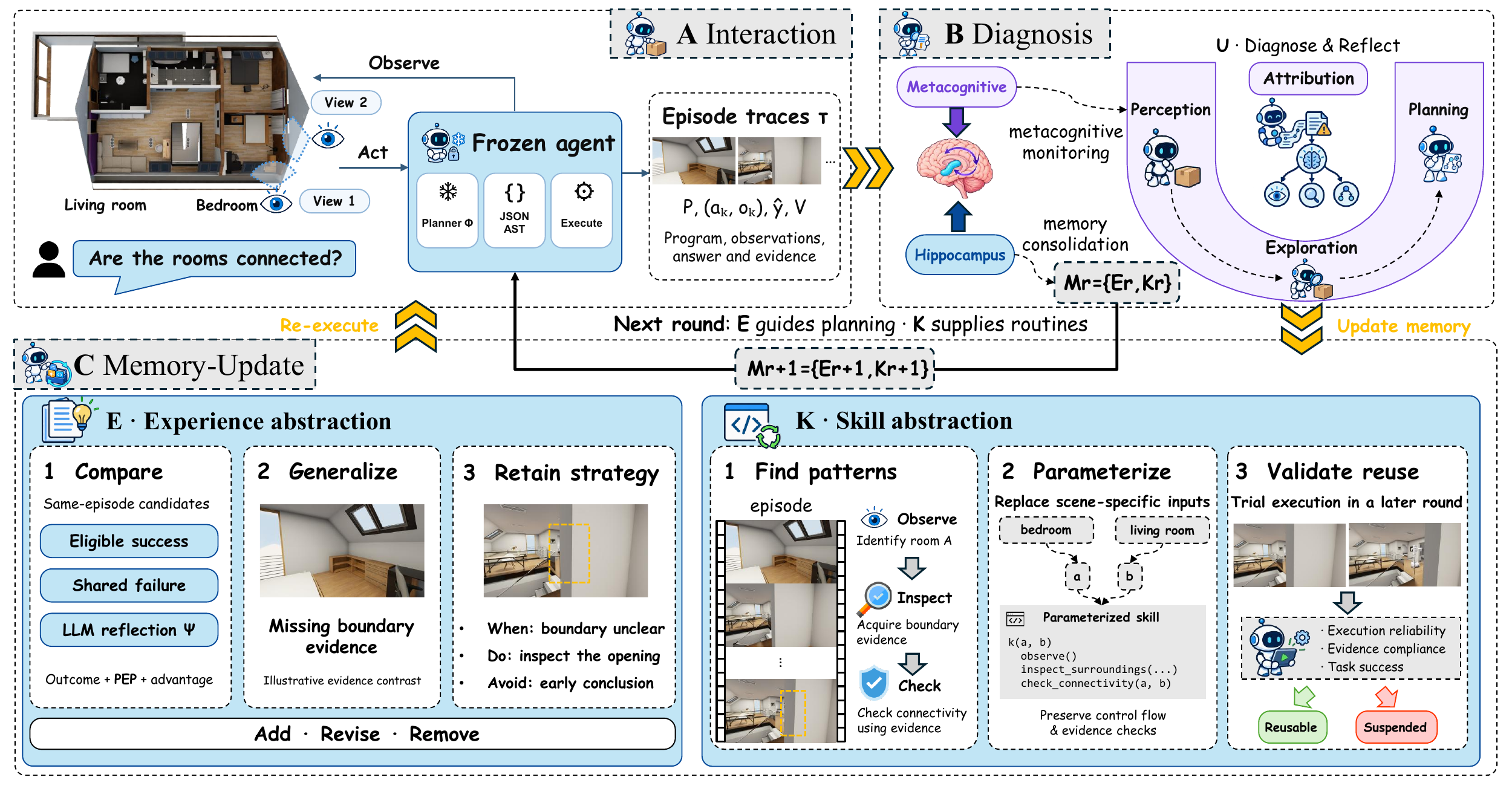}
\caption{\small
\textbf{Overview of \textsc{HiMeco}.}
\textbf{(A)} The frozen planner interacts with the environment and collects evidence.
\textbf{(B)} The PEP diagnostic module identifies perception, exploration, and planning failures.
\textbf{(C)} Strategies $\mathcal{E}$ and skills $\mathcal{K}$ are updated between rounds and used in later interactions.
}
\label{fig:method}
\end{figure*}

\paragraph{Brain-Inspired Cognitive Evolution.}
Brain-inspired approaches draw on mechanisms such as metacognitive monitoring and hippocampal memory consolidation to support adaptive behavior \citep{fleming2026integrative,huang2024replay}. Related work explores dual-system reasoning \citep{shi2026memoryvla}, cognitive maps \citep{liao2024learning}, episodic memory \citep{shu2026remem,zheng2025esceme}, and predictive world models \citep{jensen2024recurrent}. Maintaining an expanding archive of agent versions \citep{zhang2026dgm} or curating procedure memory from verified experience \citep{zhang2026sma} echoes such complementary memory systems, yet these designs select what to retain mainly by task success rather than by diagnosed failure type. These perspectives motivate linking retrospective evaluation with selective memory updates. Our framework draws on this functional analogy to couple perception, exploration, and planning diagnostics with strategy and skill consolidation.

\section{Method}

\subsection{Problem Setup}
\label{sec:problem_setup}

Figure~\ref{fig:method} summarizes the interaction, diagnosis, and memory-update process of \textsc{HiMeco}. In round $r$, a frozen planner generates programs from input $x$ and memory $\mathcal{M}_r=(\mathcal{E}_r,\mathcal{K}_r)$, comprising a strategy bank and a skill library. The diagnostic module $\mathcal{U}$ evaluates their execution trajectories $\mathcal{T}_r$ to guide the round-end update
$
\mathcal{M}_{r+1}\leftarrow\mathrm{Update}(\mathcal{M}_r,\mathcal{T}_r,\mathcal{U}).
$

Each episode has input $x=(o_0,q)$, where $o_0$ contains the initial RGB image and pose, and $q$ is a spatial query. The frozen VLM planner $\Phi$ generates $P\sim\Phi(\cdot\mid x,\mathcal{A},\mathcal{M}_r)$ using observation, navigation, and reasoning primitives $\mathcal{A}$. The program, a JSON abstract syntax tree (AST), calls primitives and branches on new observations; episode-local working memory stores observations, intermediate states, and evidence.

Execution yields $\tau=(P,\{(a_k,o_k)\}_{k=1}^{K},\hat y,V)$, where $a_k,o_k$ denote the $k$-th action and resulting observation, $K$ counts actions within budget $T_{\max}$, $\hat y$ is the answer, and $V$ records evidence checks. Given ground truth $y$, correctness $r_{\mathrm{task}}$, evidence eligibility $\ell$, and learning success $s$ are
\begin{equation}
\begin{aligned}
r_{\mathrm{task}}(\tau) &= \mathbbm{1}[\hat{y}=y],\quad
\ell(\tau) =
\begin{cases}
\mathbbm{1}[\mathrm{ContractValid}(\tau,q)],\\
1, 
\end{cases}\quad
s(\tau) = r_{\mathrm{task}}(\tau)\,\ell(\tau).
\end{aligned}
\label{eq:learning_success}
\end{equation}
Here, $\mathbbm{1}[\cdot]$ is the indicator function, equal to one when its condition holds and zero otherwise. $\mathrm{ContractValid}(\tau,q)$ indicates whether the recorded evidence passes the final contract for query $q$. $r_{\mathrm{task}}(\tau)$ measures answer correctness, $\ell(\tau)$ indicates evidence eligibility, and $s(\tau)$ indicates eligibility as a successful memory example.

\subsection{Metacognitive Diagnosis}
\label{sec:diagnosis}

After execution, $\mathcal{U}$ uses task outcomes, traces, and privileged simulator metadata to produce diagnostic scores:
\begin{equation}
\mathbf{u}(\tau)=
\big(u_{\mathrm{per}}(\tau),u_{\mathrm{exp}}(\tau),u_{\mathrm{pla}}(\tau)\big)
\in[0,1]^3.
\end{equation}
The subscripts $\mathrm{per}$, $\mathrm{exp}$, and $\mathrm{pla}$ denote perception, exploration, and planning, respectively. Scores reflect failure evidence rather than calibrated probabilities. Labels and privileged metadata are unavailable to the online planner.

\textbf{PEP diagnosis.}
An incorrect answer alone cannot distinguish faulty perception, insufficient exploration, and flawed planning, which require different corrective strategies.
\textbf{Perception} checks visual-output validity and discrepancies between symbolic results and simulator evidence, accounting for observation coverage.
\textbf{Exploration} assesses reference-viewpoint proximity, pose diversity, and missing observations.
\textbf{Planning} checks program validity, primitive execution, termination, answer return, and budget compliance; a frozen LLM auditor additionally reviews action selection and control flow.Let $\mathcal{D}=\{\mathrm{per},\mathrm{exp},\mathrm{pla}\}$ denote the diagnostic dimensions. The primary label is
\begin{equation}
f^*(\tau)=
\begin{cases}
\mathrm{none}, & \text{if correct with no planning anomaly},\\
\arg\max_{d\in\mathcal{D}} u_d(\tau), & \text{otherwise}.
\end{cases}
\end{equation}
Reflection retains all three scores to account for co-occurring failures.

\textbf{Candidate comparison.}
We compare candidates for the same episode under different memory conditions, including an experience-only reference and a trial-skill condition when available. We score each trajectory by
\begin{equation}
R(\tau) = 2r_{\mathrm{task}}(\tau) - 0.75\bar{u}(\tau) + 0.1\eta(\tau),
\end{equation}
where $\bar{u}(\tau) = \frac{1}{3}\sum_{d\in\mathcal{D}}u_d(\tau)$ and $\eta(\tau) = 1 - \min(1,n_{\mathrm{act}}(\tau)/T_{\max})$ with $n_{\mathrm{act}}(\tau)$ denoting the number of executed low-level actions. Correct trajectories score at least $1.25$ while incorrect ones score at most $0.1$, prioritizing correctness while using diagnostics and efficiency to distinguish candidates with the same correctness outcome. For the candidate group $\mathcal{C}$ of a given episode, the relative advantage
\begin{equation}
\mathrm{Adv}(\tau) = R(\tau) - \frac{1}{|\mathcal{C}|}\sum_{\tau'\in\mathcal{C}}R(\tau')
\end{equation}
guides comparative reflection alongside diagnostic feedback, without gradient updates. Positive-example admission separately requires $s(\tau)=r_{\mathrm{task}}(\tau)\ell(\tau)=1$, preventing favorable scores from bypassing evidence contracts.

\subsection{Strategy Experience Bank}
\label{sec:experience}

Individual trajectories contain instance-specific details that transfer poorly across tasks. The strategy bank $\mathcal{E}_r$ records when a strategy applies, how to acquire evidence, and which failures to avoid. Let $\mathcal{C}_r(x)$ denote the candidates for episode $x$ in round $r$. Their diagnostic information forms $\mathcal{B}_r(x) = \{z(\tau):\tau\in\mathcal{C}_r(x)\}$, where $z(\tau)$ summarizes execution, PEP scores, the primary diagnostic label, relative advantage, evidence eligibility, and learning success. Updates come from three sources. When the eligible-success set
$
\mathcal{C}_r^+(x)=\{\tau\in\mathcal{C}_r(x):s(\tau)=1\}
$
is nonempty, we select the candidate with the highest advantage, breaking ties by fewer actions. We abstract its use of primitives into a reusable strategy, excluding answer values and episode-specific coordinates.

Relative advantages provide limited guidance when candidates share the same failure. We therefore flag candidate collapse when
$
|\mathcal{C}_r(x)|\ge 2,\quad
\sum_{\tau\in\mathcal{C}_r(x)}s(\tau)=0,
$
and all candidates return the same answer. Deterministic rules produce corrective guidance, such as acquiring another viewpoint or adding a visual check, annotated with the primary diagnostic label.

An LLM reflector $\Psi$ compares candidates within each bundle to identify useful strategies and lessons from failures. Its updates,
$
\Delta\mathcal{E}_r^{\mathrm{refl}}=\Psi(\{\mathcal{B}_r(x)\}_x),
$
are combined with updates from selected successful candidates
$
\Delta\mathcal{E}_r^{\mathrm{champ}}
$
and shared failures
$
\Delta\mathcal{E}_r^{\mathrm{fail}}:
$
\begin{equation}
\mathcal{E}_{r+1}
=
\mathrm{Apply}\!\left(
\mathcal{E}_r,
\Delta\mathcal{E}_r^{\mathrm{champ}},
\Delta\mathcal{E}_r^{\mathrm{fail}},
\Delta\mathcal{E}_r^{\mathrm{refl}}
\right).
\end{equation}
Here, $\mathrm{Apply}$ adds, revises, or removes entries at the round boundary under the memory capacity limit. The strategy bank may retain episode-specific entries during adaptation, while final-round strategy retrieval uses only task-level guidance.

\subsection{Composite Skills}
\label{sec:skills}

The skill library $\mathcal{K}$ stores reusable routines for acquiring evidence. Since recurring patterns may still be unreliable, abstraction is followed by execution-based validation. Skills replace object-specific inputs with parameters, retain control flow and evidence checks, and exclude answer submission. For a program pattern $p$, let $\mathcal{X}_p$ contain episodes with occurrences eligible for abstraction, and $\mathcal{X}_p^+\subseteq\mathcal{X}_p$ those with eligible occurrences in successful trajectories ($s(\tau)=1$). To become a candidate skill, a pattern must satisfy
\begin{equation}
|\mathcal{X}_p| \ge m,
\qquad
\frac{|\mathcal{X}_p^+|}{|\mathcal{X}_p|}
\ge\theta_{\mathrm{syn}},
\end{equation}
where $m\ge1$ is the minimum episode count and $\theta_{\mathrm{syn}}\in[0,1]$ the required success rate. Each episode counts once per set. Systematic failures also yield corrective routines, such as entity verification and viewpoint changes.

Candidate skills undergo trial execution before regular reuse; unreliable skills can be suspended. For each skill $k$, let $N_{\mathrm{use}},N_{\mathrm{exec}},N_{\mathrm{valid}},N_{\mathrm{succ}}$ count episodes with invocation, valid execution, contract-valid execution, and successful contract-valid execution, respectively. We smooth these rates using $\rho(a,b)=(a+\lambda\mu)/(b+\lambda)$ to reduce sensitivity to sparse data, where $\mu\in[0,1]$ is the prior mean and $\lambda>0$ is the prior strength. Skill quality is
\begin{equation}
\begin{aligned}
Q(k)=&w_s\rho(N_{\mathrm{succ}},N_{\mathrm{valid}})
+w_c\rho(N_{\mathrm{valid}},N_{\mathrm{exec}})\\
&+w_e\rho(N_{\mathrm{exec}},N_{\mathrm{use}})+w_a\tilde{\eta}(k),
\end{aligned}
\end{equation}
where the nonnegative weights sum to one and $\tilde{\eta}(k)$ is the prior-smoothed median action efficiency over successful contract-valid invocations. Validation checks execution reliability, evidence compliance, and task success to determine reuse eligibility; retrieval prioritizes lifecycle status, empirical success, and episode support.

\subsection{Interaction and Evolution Protocol}
\label{sec:evolution_protocol}

Each round uses a fixed memory snapshot. For each episode, candidates are evaluated under an experience-only condition and, when available, a trial-skill condition. After all candidates finish, PEP diagnostics guide strategy updates, while execution outcomes update skill support and reuse eligibility. The updates take effect in the next round, allowing experience from one episode to guide other episodes in the same subtask. After $R$ rounds, the memory is frozen, and only validated skills are reused.

\section{Experiments}

\subsection{Experimental Setup}
\label{sec:setup}

\textbf{Datasets.}
We evaluate active and passive embodied spatial reasoning on ESI-Bench \citep{hong2026esibench} and EmbSpatial-Bench \citep{du2024embspatial}, respectively. ESI-Bench, built on OmniGibson, assesses spatial reasoning through interaction with the environment. We evaluate repeated-episode adaptation on 208 fixed episodes spanning nine tasks, with identical episodes and initial scenes across our settings. Cross-episode transfer is examined separately through disjoint source-target examples. For EmbSpatial-Bench, we randomly sample 300 questions per spatial relation category, totaling 1,800 questions across six categories. Its fixed visual inputs lack the interaction feedback required for PEP failure attribution. We therefore evaluate \textsc{HiMeco} without evolution against API inference using the same backbone and inputs. Sampling details are provided in the appendix.

\textbf{Models.}
We use GPT-5.6-Terra and Gemini-3-Flash with frozen parameters and shared task inputs, tool interfaces, and scoring criteria. On ESI-Bench, we also evaluate a \textit{Passive Single-View} baseline using only the initial image. Experiments run on a server with eight NVIDIA RTX A6000 GPUs, accessing VLMs through APIs. We retain execution traces for analysis and provide prompts, tool configurations, and hyperparameters in the appendix.

\textbf{Evaluation Metrics.}
We report answer accuracy (\%), averaged equally across the nine ESI-Bench tasks and across all questions on EmbSpatial-Bench. On ESI-Bench, we additionally report mean per-episode wall-clock time, including diagnosis but excluding between-round reflection and memory updates. Diagnostic analysis uses mean perception, exploration, and planning failure scores in $[0,1]$ (lower is better) and primary failure-label counts normalized by all evaluated episodes. We report absolute score changes and relative runtime changes from Round-0 to the final round, alongside action counts in cross-episode generalization examples.

\subsection{Results}
\label{sec:results}

\paragraph{Results on ESI-Bench.}
Table~\ref{tab:main_results} compares \textsc{HiMeco} with existing methods reporting ESI-Bench results, covering VLM baselines, interactive agents like ParallelWorld \citep{chen2026parallelworld}, and self-evolving agents such as SHAPER \citep{wang2026shaper}. \textsc{HiMeco} achieves state-of-the-art average accuracy of 69.19\% on the evaluated ESI-Bench tasks. Relative to the corresponding active baselines, accuracy improves from 56.89\% to 69.19\% with GPT-5.6 and from 54.74\% to 62.51\% with Gemini-3-Flash, yielding gains of 12.30 and 7.77 percentage points without task-level degradation. The largest gains occur on Spatial Distance (+25.0 points) and Physical Contact (+24.0) with GPT-5.6, and Spatial Distance (+15.0) and Connectivity (+11.1) with Gemini-3-Flash. These improvements demonstrate the effectiveness of diagnosis-guided evolution for active embodied spatial reasoning, although Counting w/ Occlusion remains challenging.

\begin{table}[t]
\centering
\small
\setlength{\tabcolsep}{3pt}
\renewcommand{\arraystretch}{1.15}
\caption{\small Accuracy (\%) on ESI-Bench.}
\label{tab:main_results}
\resizebox{\textwidth}{!}{%
\begin{tabular}{@{}ll*{10}{c}@{}}
\toprule
\multicolumn{2}{@{}l}{Backbone / Method}
& \shortstack{Metric\\Comparison}
& \shortstack{Cognitive\\Mapping}
& \shortstack{Perceptual\\Grounding}
& \shortstack{Action\\Sequencing}
& \shortstack{Spatial\\Relations}
& \shortstack{Specular\\Reflection}
& \shortstack{Enumerative\\Perception}
& \shortstack{Temporal\\Understanding}
& \shortstack{Physical\\Dynamics}
& Avg. \\
\midrule
\multicolumn{12}{@{}l}{\textit{Our evaluation}} \\
\multicolumn{2}{@{}l}{Selected subtask}
& \shortstack{Spatial\\Distance}
& Connectivity
& \shortstack{Partial\\Occlusion}
& \shortstack{Action Order\\Inference}
& \shortstack{Physical\\Contact}
& Correspondence
& \shortstack{Counting w/\\Occlusion}
& \shortstack{Unobserved\\Change}
& \shortstack{Inclined\\Plane}
& - \\
\multicolumn{2}{c}{$n$}
& 20 & 27 & 20 & 20 & 25 & 20 & 20 & 29 & 27 & - \\
\addlinespace[4pt]
\multirow{3}{*}{GPT-5.6}
& Passive
& 45.0 & 59.3 & 45.0 & 35.0 & 68.0 & 65.0 & 10.0 & 34.5 & 37.0 & 44.31 \\
& Active
& 60.0 & 70.4 & 50.0 & 40.0 & 64.0 & 75.0 & 10.0 & 75.9 & 66.7 & 56.89 \\
\rowcolor{hmecobg}
& \textsc{HiMeco}
& \textbf{85.0} & \textbf{74.1} & \textbf{65.0} & \textbf{55.0} & \textbf{88.0} & \textbf{80.0} & \textbf{15.0} & \textbf{82.8} & \textbf{77.8} & \textbf{69.19} \\
\addlinespace[5pt]
\multirow{3}{*}{Gemini-3-Flash}
& Passive
& 70.0 & 59.3 & 35.0 & 5.0 & 68.0 & 20.0 & 5.0 & 41.4 & 51.9 & 39.51 \\
& Active
& 75.0 & 66.7 & \textbf{65.0} & 25.0 & 76.0 & 30.0 & 5.0 & 75.9 & 74.1 & 54.74 \\
\rowcolor{hmecobg}
& \textsc{HiMeco}
& \textbf{90.0} & \textbf{77.8} & \textbf{65.0} & \textbf{35.0} & \textbf{84.0} & \textbf{40.0} & \textbf{10.0} & \textbf{79.3} & \textbf{81.5} & \textbf{62.51} \\
\midrule
\multirow{2}{*}{GPT-5$^\ast$}
& Passive
& 53.9 & 68.3 & 32.6 & 36.4 & 40.0 & 39.8 & 3.3 & 40.5 & 57.4 & 41.36 \\
& Active
& 58.6 & 68.3 & 47.4 & 41.6 & 64.2 & 52.3 & 13.3 & 51.4 & 77.0 & 52.68 \\
\addlinespace[4pt]
\multirow{2}{*}{Gemini-3.1$^\ast$}
& Passive
& 52.6 & 51.7 & 30.5 & 44.2 & 60.8 & 37.5 & 3.3 & 37.2 & 65.6 & 42.60 \\
& Active
& 59.9 & 60.0 & 70.5 & 54.5 & 70.0 & 42.0 & 10.0 & 47.3 & 83.6 & 55.31 \\
\midrule
\addlinespace[4pt]
\multicolumn{2}{@{}l}{ParallelWorld$^{\dagger}$}
& 67.11 & 60.00 & 77.89 & 61.04 & 71.43 & 48.86 & 30.00 & 89.86 & 85.25 & 65.72 \\
\midrule
\multicolumn{2}{@{}l}{$n$}
& 21 & 17 & 49 & 5 & 51 & 20 & 18 & 19 & 10 & - \\
\multicolumn{2}{@{}l}{SHAPER$^{\ddagger}$}
& 47.6 & 23.5 & 69.4 & 20.0 & 54.9 & 60.0 & 27.8 & 47.4 & 40.0 & 42.9 \\
\bottomrule
\end{tabular}%
}
\end{table}

\begin{figure}[t]
\centering
\includegraphics[width=\textwidth,height=0.6\textheight,keepaspectratio]{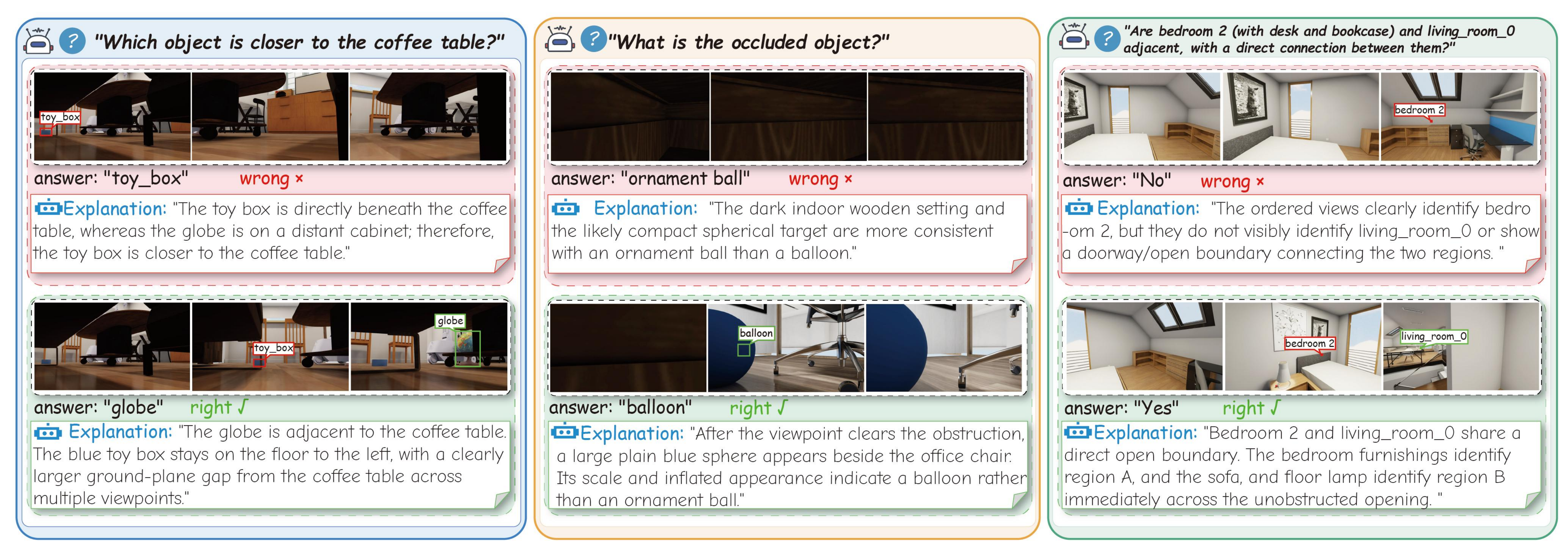}
\caption{
\textbf{Qualitative examples of embodied spatial reasoning.}
Incorrect and correct predictions are shown with their observations and explanations.
}
\label{fig:qualitative}
\end{figure}

\paragraph{Qualitative Analysis.}
Figure~\ref{fig:qualitative} presents three representative cases. In Spatial Distance, the agent without memory selects the toy box from incomplete cross-view evidence, while \textsc{HiMeco} tracks both objects across viewpoints and correctly selects the globe. In Partial Occlusion, it changes the viewpoint to reveal the hidden object, correcting ornament ball to balloon. In Connectivity, it explores the room boundary and finds the direct opening between bedroom-2 and living room-0, correcting No to Yes. These examples show that diagnosis-guided memory helps resolve spatial ambiguity, occlusion, and incomplete exploration.

\begin{figure}[t]
\centering
\includegraphics[width=\textwidth,height=0.65\textheight,keepaspectratio]{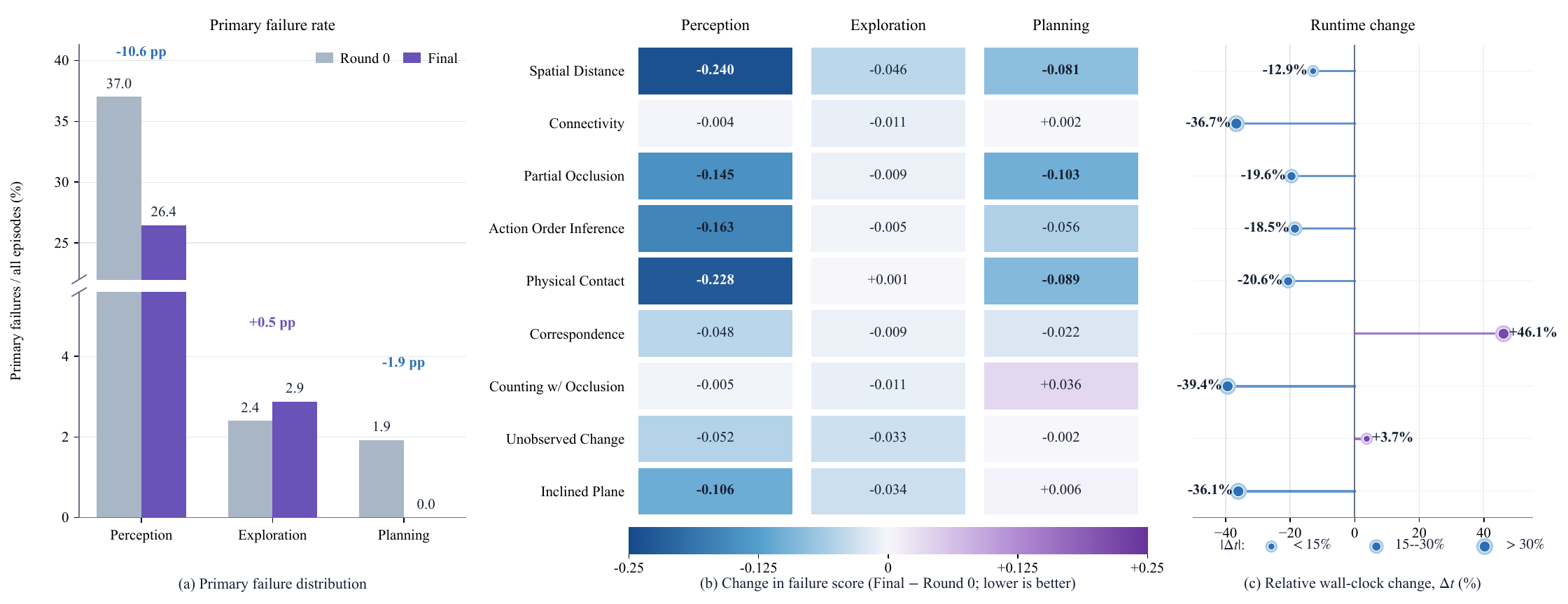}
\caption{
\textbf{Failure diagnosis and execution efficiency on ESI-Bench.}
Round-0 versus the final evolution round:
\textbf{(a)} Primary failure counts normalized by all evaluation episodes.
\textbf{(b)} Changes in mean diagnostic scores.
\textbf{(c)} Relative changes in mean wall-clock time per episode.
Negative values in (b) and (c) indicate reductions.
}
\label{fig:failure}
\end{figure}

\paragraph{Failure Diagnosis and Efficiency.}
Figure~\ref{fig:failure} shows that \textsc{HiMeco} improves accuracy while reducing diagnostic failure scores and execution time. From Round-0 to the final round, perception scores decrease across all nine tasks, with the largest reductions on Spatial Distance ($-0.240$) and Physical Contact ($-0.228$), accompanying accuracy gains of 25 and 24 percentage points. Exploration scores decrease on eight tasks, while planning scores decrease on six. Across all evaluation episodes, perception-primary failures fall from 37.0\% to 26.4\%, and planning-primary failures fall from 1.9\% to zero. The latter indicates a change in dominant failure attribution, rather than the absence of all planning anomalies. These improvements generally require less execution time. Mean wall-clock time decreases on seven of nine tasks by 12.9--39.4\%, with a 14.9\% reduction when averaged equally across tasks. Correspondence and Unobserved Change instead take 46.1\% and 3.7\% longer, respectively. Overall, the results support the value of consolidated strategies and skills in improving execution reliability and efficiency.

\begin{figure}[t]
\centering
\includegraphics[width=1.0\textwidth,height=0.5\textheight,keepaspectratio]{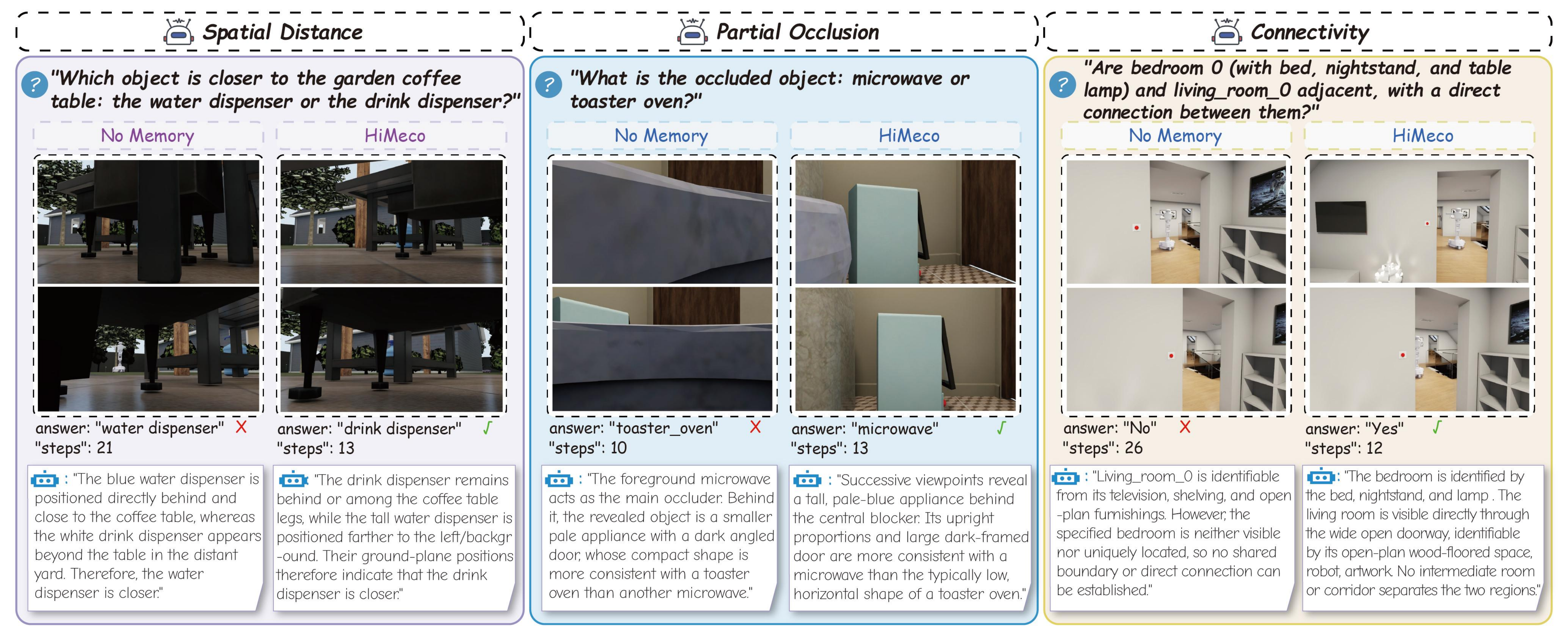}
\caption{
Cross-episode generalization.
}
\label{fig:cross}
\end{figure}

\paragraph{Cross-Episode Generalization.}
The main results evaluate adaptation on repeated episodes. To examine transfer, each case here applies a single learned experience to an unseen episode within the same subtask. Source and target episodes are disjoint, and memory remains frozen during evaluation. Figure~\ref{fig:cross} shows \textsc{HiMeco} succeeding where the memory-free agent fails. Comparing ground-plane positions across views in Spatial Distance, revealing an occluded microwave in Partial Occlusion, and verifying a direct doorway connection in Connectivity. Action counts change from 21 to 13, 10 to 13, and 26 to 12, respectively, illustrating both more efficient exploration and additional evidence gathering when needed. These cases provide qualitative evidence that individual learned experiences can support evidence acquisition and spatial reasoning on unseen episodes.

\begin{table}[t]
\centering
\scriptsize
\setlength{\belowcaptionskip}{6pt}
\setlength{\tabcolsep}{3pt}
\renewcommand{\arraystretch}{1.05}
\caption{\small Component ablations. The diagnostic ablation replaces structured PEP-guided reflection with generic reflection while retaining general execution reflection and program-change suggestions.}
\label{tab:ablation_components}
\begin{tabular}{cccccccc}
\toprule
$\mathcal{E}$ & $\mathcal{K}$ & $\mathcal{U}$ & Evol.
& Spatial Distance
& Action Order Inference
& Physical Contact
& Inclined Plane \\
\midrule
& & & &
60 & 40 & 64 & 66.7 \\
\checkmark & & \checkmark & \checkmark &
65 & 55 & 80 & 74.1 \\
& \checkmark & \checkmark & \checkmark &
70 & 45 & 76 & 70.4 \\
\checkmark & \checkmark & & \checkmark &
65 & 45 & 80 & 74.1 \\
\checkmark & \checkmark & \checkmark & \checkmark &
\textbf{85} & \textbf{55} & \textbf{88} & \textbf{77.8} \\
\bottomrule
\end{tabular}
\end{table}

\paragraph{Ablation Study.}
Table~\ref{tab:ablation_components} shows that full \textsc{HiMeco} achieves the best or tied-best accuracy across all four tasks. Removing the strategy bank $\mathcal{E}$ decreases accuracy by 7.4--15.0 percentage points, while removing the skill library $\mathcal{K}$ reduces accuracy on three tasks, with the largest drop on Spatial Distance (20.0 points). Replacing structured PEP-guided reflection with generic reflection reduces accuracy by 3.7--20.0 percentage points across the four tasks. The generic reflector uses execution outcomes to propose general lessons and program changes, without explicit perception, exploration, or planning labels or scores as reflection inputs.

\begin{table}[t]
\centering
\small
\setlength{\tabcolsep}{3pt}
\renewcommand{\arraystretch}{1.16}
\caption{Accuracy (\%) on EmbSpatial-Bench.}
\label{tab:embspatial_results}
\begin{tabular*}{\linewidth}{@{\extracolsep{\fill}}lcc@{}}
\toprule
Relation & Gemini-3-Flash & \shortstack{HiMeco\\w/o Evolution} \\
\midrule
Left & 95.33 & 92.33 \\
Right & 84.67 & 82.67 \\
Above & 75.33 & 88.00 \\
Under & 73.67 & 85.00 \\
Far & 57.33 & 65.00 \\
Close & 67.33 & 64.33 \\
\midrule
Overall & 75.61 & \textbf{79.56} \\
\bottomrule
\end{tabular*}
\end{table}

\paragraph{Results on EmbSpatial-Bench.}
We further evaluate \textsc{HiMeco} on passive embodied spatial reasoning, where fixed observations preclude active evidence acquisition and the interaction feedback needed for full PEP failure attribution is unavailable. Without evolution, \textsc{HiMeco} improves accuracy from 75.61\% to 79.56\% over direct Gemini-3-Flash API inference, a gain of 3.94 percentage points (Table~\ref{tab:embspatial_results}). Gains are largest on Above (+12.67 points), Under (+11.33), and Far (+7.67), while Left, Right, and Close decrease by 2.0--3.0 points. These results show that the static reasoning configuration remains effective when diagnosis-guided evolution is unavailable.

\section{Conclusion}

This paper presented \textsc{HiMeco}, a brain-inspired self-evolving agent for embodied spatial intelligence. Its Reflection mechanism couples PEP failure diagnosis module with the consolidation of reusable strategies and executable skills. Evidence contracts and execution-based validation govern skill retention and reuse, enabling adaptation without updating model parameters. \textsc{HiMeco} achieves SOTA performance on ESI-Bench and improves accuracy on EmbSpatial-Bench by 3.94 percentage points without evolution.

\subsection*{AI Use Statement}

In this paper, we used generative AI tools to polish the manuscript's language refinement, grammar checking, little coding editing, and improve style figures. All AI-assisted text was reviewed by the authors.

\subsection*{Reproducibility Statement}

The main text describes our framework and evolution procedure. The appendix provides detailed method specifications, dataset preprocessing and sampling procedures, prompts, tool configurations, and hyperparameters. We will publicly release the implementation and evaluation code to support reproduction of all proposed components and experiments.

\subsection*{Author Contributions}

The author team jointly contributed to the conceptualization, development, and evaluation of this work. Contributions included methodology development, experimental design and execution, analysis of results, preparation of methodological illustrations and experimental figures, manuscript drafting and revision, reference curation, and manuscript consistency checks. The team also contributed research supervision, scientific advice, project management, and overall coordination. Equal-contribution designations and corresponding-author information are indicated in the author list above.

\bibliographystyle{plainnat}
\bibliography{iclr2027_conference}

\clearpage
\appendix
\input{appendix921}

\end{document}

%% file: appendix921.tex

\hypersetup{hidelinks}
\raggedbottom

\section*{Appendix contents}

This appendix supplements the main-text experiments with evaluation
protocols, evidence checks, PEP diagnosis, memory updates, and prompts.

\begingroup
\small
\newcommand{\appentry}[2]{\noindent\hyperref[#1]{\ref*{#1}\quad #2}\dotfill\pageref{#1}\par}

\appentry{app:qualitative}{Additional qualitative examples}
\appentry{app:algorithms}{Algorithm overview ($\mathcal E$, $\mathcal K$, and $\mathcal U$)}
\appentry{app:protocol}{Evaluation protocol and candidate comparison}
\appentry{app:data}{Datasets, sampling, and episode lists}
\appentry{app:execution}{Program execution and evidence checks}
\appentry{app:diagnosis}{Perception, exploration, and planning diagnosis}
\appentry{app:memory}{Strategy and skill updates}
\appentry{app:implementation}{Implementation settings}
\appentry{app:prompt_templates}{Prompt templates}
\endgroup

\section{Additional Qualitative Examples}
\label{app:qualitative}

\begin{figure}[H]
\centering
\includegraphics[width=\textwidth,height=0.49\textheight,keepaspectratio]{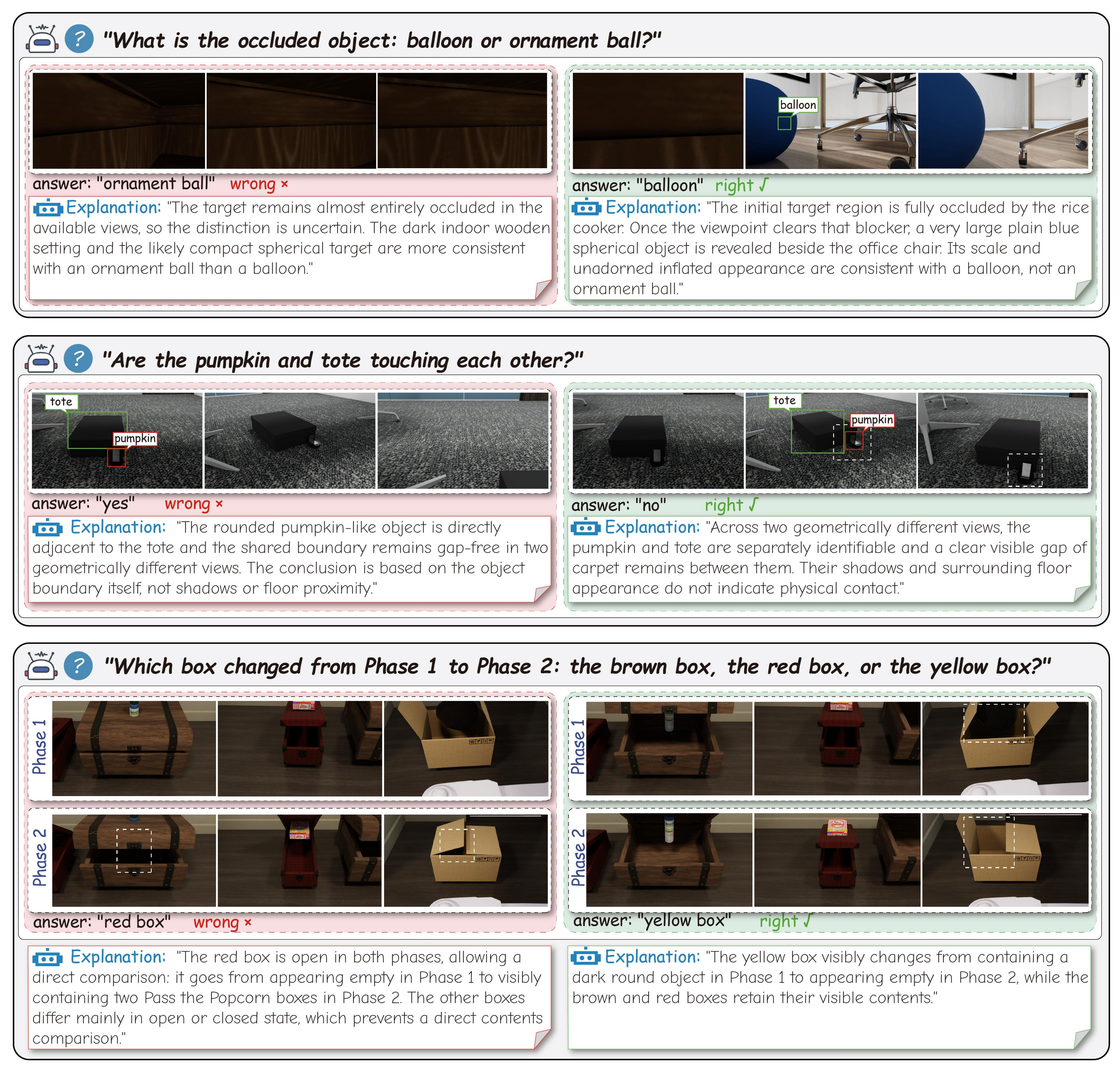}
\caption{Additional qualitative examples of embodied spatial reasoning.
Incorrect and correct predictions are presented with their observations
and explanations.}
\label{fig2}
\end{figure}

Figure~\ref{fig2} provides additional examples with observations,
predictions, and explanations. The examples show how observations support the returned answers.

\clearpage

\section{Algorithm Overview}
\label{app:algorithms}

Algorithm~\ref{alg:evolution_detailed} gives the four-round schedule.
Algorithms~\ref{alg:strategy_update} and~\ref{alg:skill_update} update the
strategy bank $\mathcal E$ and skill library $\mathcal K$.
Section~\ref{app:diagnosis} defines diagnosis $\mathcal U$, which runs
once per completed trajectory.

We use the main-text notation $\mathcal M_r=(\mathcal E_r,\mathcal K_r)$,
$\mathcal C_r(x)$ for candidate trajectories, $\mathcal T_r$ for their union,
and $\mathcal B_r(x)$ for a diagnostic bundle. For the reported schedule,
$r=t+1$, where $t=0,1,2,3$ is the displayed execution round. Thus,
$\mathcal M_1$ is empty and $\mathcal M_4$ is frozen for Round~3.
Here $c_r$ is the candidate count. In the reported evaluation, the planner
generates each candidate from the round's memory snapshot.

\begin{algorithm}[H]
\small
\caption{Four-round execution and memory updates}
\label{alg:evolution_detailed}
\begin{algorithmic}[1]
\Require Episodes $\mathcal D$; frozen planner $\Phi$ and reflector $\Psi$;
diagnosis $\mathcal U$; task budgets $T_{\max}(q)$
\Ensure Round-0 and Round-3 results; frozen memory $\mathcal M_4$
\State $\mathcal M_1\gets(\varnothing,\varnothing)$
\For{$r=1,2,3,4$}
    \State Fix $\mathcal M_r$ for this round; $\mathcal T_r\gets\varnothing$
    \State $c_r\gets1$ if $r\in\{1,4\}$, otherwise $3$
    \For{episode $x=(o_0,q)\in\mathcal D$}
        \State $\mathcal C_r(x)\gets\varnothing$
        \For{$i=0,\ldots,c_r-1$}
            \State Select candidate memory and eligible skills (Section~\ref{app:candidates})
            \State Generate and validate $P$ using $\Phi$ and the selected memory
            \State Reset $x$; execute $P$ within $T_{\max}(q)$ to obtain $\tau$
            \State Evaluate $r_{\mathrm{task}}(\tau)$, $\ell(\tau)$, and $s(\tau)$ (Section~\ref{app:contracts})
            \State Compute diagnostic scores and $f^*(\tau)$ with $\mathcal U$ (Section~\ref{app:diagnosis})
            \State $\mathcal C_r(x)\gets\mathcal C_r(x)\cup\{\tau\}$
        \EndFor
        \State Compute $R(\tau)$ and $\mathrm{Adv}(\tau)$ within $\mathcal C_r(x)$ (Section~\ref{app:reward})
        \State Form $\mathcal B_r(x)$; $\mathcal T_r\gets\mathcal T_r\cup\mathcal C_r(x)$
    \EndFor
    \If{the round is incomplete}
        \State Stop without advancing the memory checkpoint
    \EndIf
    \If{$r<4$}
        \State Update $\mathcal E_r$ using Algorithm~\ref{alg:strategy_update}
        \State Update $\mathcal K_r$ using Algorithm~\ref{alg:skill_update}
        \State Save $\mathcal M_{r+1}\gets(\mathcal E_{r+1},\mathcal K_{r+1})$
    \EndIf
\EndFor
\end{algorithmic}
\end{algorithm}

\clearpage

\subsection{Strategy-bank update ($\mathcal E$)}
\label{app:strategy_algorithm}

This update combines a successful candidate's procedure, a correction for
shared failures, and the reflector's suggestions. It uses the diagnostic
bundles already collected in the round; it does not rerun diagnosis.

\begin{algorithm}[H]
\small
\caption{Strategy-bank update ($\mathcal E$)}
\label{alg:strategy_update}
\begin{algorithmic}[1]
\Require $\mathcal E_r$; candidate groups $\mathcal C_r(x)$ and bundles $\mathcal B_r(x)$; frozen reflector $\Psi$
\State $\Delta\mathcal E_r^{\mathrm{champ}}\gets\varnothing$; $\Delta\mathcal E_r^{\mathrm{fail}}\gets\varnothing$
\For{each episode $x$}
    \State $\mathcal C_r^+(x)\gets\{\tau\in\mathcal C_r(x):s(\tau)=1\}$
    \If{$\mathcal C_r^+(x)\ne\varnothing$}
        \State Select the highest-advantage success; break ties by fewer actions
        \State Add its evidence-acquisition strategy to $\Delta\mathcal E_r^{\mathrm{champ}}$
    \ElsIf{$|\mathcal C_r(x)|\ge2$ and all candidates return the same answer}
        \State Add a diagnosis-guided correction to $\Delta\mathcal E_r^{\mathrm{fail}}$
    \EndIf
\EndFor
\State $\Delta\mathcal E_r^{\mathrm{refl}}\gets\Psi(\{\mathcal B_r(x)\}_x)$
\State Normalize update fields and supporting candidate references
\State Apply the three sets of updates to $\mathcal E_r$ to obtain $\mathcal E_{r+1}$
\State Compact entries and enforce the task capacity (Section~\ref{app:strategy})
\State \Return $\mathcal E_{r+1}$
\end{algorithmic}
\end{algorithm}

\subsection{Skill-library update ($\mathcal K$)}
\label{app:skill_algorithm}

Existing skills are assessed from actual calls. New skills enter a trial
state for later rounds. The sets $\mathcal X_p$ and $\mathcal X_p^+$
contain episodes with a structurally eligible occurrence of pattern $p$
and with at least one such occurrence in a learning-success trajectory,
respectively. Each episode counts once per set in an update.

\begin{algorithm}[H]
\small
\caption{Skill-library update ($\mathcal K$)}
\label{alg:skill_update}
\begin{algorithmic}[1]
\Require $\mathcal K_r$; completed trajectories $\mathcal T_r$; minimum support $m$; success threshold $\theta_{\mathrm{syn}}$
\State $\mathcal K_{r+1}\gets$ a working copy of $\mathcal K_r$
\For{each skill $k$ called in $\mathcal T_r$}
    \State Update $N_{\mathrm{use}},N_{\mathrm{exec}},N_{\mathrm{valid}},N_{\mathrm{succ}}$ from episode-level sets
    \State Update its status using current-round outcomes (Section~\ref{app:lifecycle})
    \State Recompute $Q(k)$ from accumulated counts (Section~\ref{app:skill_quality})
\EndFor
\For{each normalized program pattern $p$ in $\mathcal T_r$}
    \State Form the episode sets $\mathcal X_p$ and $\mathcal X_p^+$
    \If{$|\mathcal X_p|\ge m$ and $|\mathcal X_p^+|/|\mathcal X_p|\ge\theta_{\mathrm{syn}}$}
        \State Parameterize inputs, keep dependencies, and remove answer submission
        \State Add the routine as a trial skill if its static checks pass
    \EndIf
\EndFor
\State Add eligible navigation patterns and corrective routines (Section~\ref{app:skill_abstraction})
\State Merge admitted routines and enforce the library capacity
\State \Return $\mathcal K_{r+1}$; new skills become available in the next round
\end{algorithmic}
\end{algorithm}

\clearpage

\section{Evaluation Protocol and Candidate Comparison}
\label{app:protocol}

\subsection{Comparison settings}
\label{app:settings}

Our ESI-Bench settings use the same episode list, initial scene
specifications, and answer-scoring rules, with frozen model parameters.
\emph{Passive Single-View} answers from the initial image and question.
\emph{Active Exploration} is the program agent at Round~0, before memory
updates. \textsc{HiMeco} updates strategies and skills between rounds
and generates programs using the resulting memory.

The main-text SOTA comparison refers to the nine ESI-Bench task categories
in Table~\ref{tab:main_results}, where \textsc{HiMeco} with GPT-5.6 has
the highest listed average accuracy, $69.19\%$. The standalone GPT-5 and
Gemini-3.1 reference rows use full-dataset results from
\citet{hong2026esibench}. ParallelWorld and SHAPER are the comparison
methods identified in the main text; the SHAPER row reports its task-wise
sample counts. Our settings use the episode lists in
Table~\ref{tab:app_episode_ids}.

\subsection{Evaluation episodes and round schedule}
\label{app:rounds}

The main ESI-Bench experiment measures adaptation on the same selected
episodes across four rounds. Each candidate is one complete program
execution from the initial scene state; strategy and skill memories carry
over between rounds. Cross-episode generalization uses the separate
procedure in Section~\ref{app:cross_episode}.

We report execution rounds as $t=0,1,2,3$ and use the main-text memory
index $r=t+1$. The snapshot before execution is therefore $\mathcal M_{t+1}$.
Updates follow rounds $0$, $1$, and $2$; Round~3 uses the frozen
$\mathcal M_4$ without an update:
\begin{equation}
\mathcal M_{r+1}
=\mathrm{Update}\!\left(\mathcal M_r,\mathcal T_r,\mathcal U\right),
\qquad r\in\{1,2,3\}.
\label{eq:app_round_update}
\end{equation}

\begin{table}[htbp]
\centering
\small
\caption{Execution schedule of the full repeated-instance method. Every
candidate in a row uses the same memory fixed before the round.}
\label{tab:app_round_schedule}
\begin{tabular}{@{}clcp{0.38\linewidth}@{}}
\toprule
Round & Candidates per episode & Update afterward & Purpose \\
\midrule
0 & 1 & Yes & Initial active baseline and first consolidation \\
1 & 3 & Yes & Candidate comparison and skill trials \\
2 & 3 & Yes & Candidate comparison and skill validation \\
3 & 1 & No & Final execution with frozen memory \\
\bottomrule
\end{tabular}
\end{table}

The schedule uses eight executions per episode, or $1{,}664$ planned
executions for $208$ episodes, excluding retries caused by simulator failures. Final
accuracy is measured from one execution per episode after adaptation.
All workers use the same memory snapshot, and updates begin only after
the round is complete.

\subsection{Candidate programs for comparison}
\label{app:candidates}

In Rounds~1 and~2, the agent generates three programs for the same question
under different memory conditions. They explore alternative strategies and
execution paths, giving the reflector concrete attempts to compare.
In the two intermediate rounds, candidate~0 receives retrieved strategies
and base primitives, candidate~1 receives strategies plus one trial skill
when available, and candidate~2 receives base primitives without either
memory. Candidate-specific instructions encourage different ways to gather
evidence; distinct paths are encouraged but not guaranteed.
If no skill is eligible, candidate~1 uses the strategy bank without a skill.
These candidates support reflection and skill validation. The final score
uses one fresh execution per episode, with a newly generated program
conditioned on the frozen memory.

Trial selection first chooses the highest-ranked active or validated skill.
If none is available, it uses a deterministic assignment from the trial
catalog. The planner either invokes the supplied skill or records why it
is not applicable; logs track actual invocations.

\subsection{Information available at each stage}
\label{app:information}

\paragraph{Program generation and execution.}
The planner receives the question, answer options, permitted RGB evidence,
the public primitive catalog, and the memory selected for that candidate.
Observation records may include the agent's own camera pose and action
history. The interpreter passes tool outputs through named variables. Ground-truth answer labels and the verifier's hidden simulator
snapshot are not supplied as online planning inputs.

Unobserved Change provides reference images from two phases and box
identities. Its reference-only mode can answer without additional movement.
Inclined Plane uses frames captured during the physical action.

\paragraph{Retrospective evaluation and diagnosis.}
After execution, the benchmark scorer evaluates the submitted answer. The
verifier uses the completed trace and a simulator snapshot containing
reference camera poses, observed camera poses, task postprocessing metadata,
and scoring information. The diagnostic rules use reference camera positions and the number of
target objects spawned by the simulator, among other fields.
The snapshot is not itself an online observation.

\paragraph{Memory consolidation.}
The updater receives candidate outcomes, their predicted answers, diagnostic
scores, execution summaries, and relative advantages. Reflection is
instructed to retain evidence-acquisition procedures and corrective lessons,
and to avoid copying ground-truth answers or episode-specific coordinates.
Answer feedback is used for memory updates, while model parameters remain
frozen.

\subsection{Cross-episode generalization}
\label{app:cross_episode}

Following the main-text experiment, each case transfers a single learned
experience to an unseen episode within the same subtask. Source and target
episodes are disjoint, and memory is frozen during target evaluation.
Figure~\ref{fig:cross} presents examples for Spatial Distance, Partial
Occlusion, and Connectivity, with action counts changing from $21$ to $13$,
$10$ to $13$, and $26$ to $12$, respectively. These cases illustrate how
learned experiences guide evidence acquisition and spatial reasoning on
unseen episodes.

\section{Datasets, Sampling, and Episode Lists}
\label{app:data}

\subsection{ESI-Bench subset and sampling}
\label{app:esi_subset}

ESI-Bench evaluation uses the same fixed subset of $208$ episodes across
nine tasks for our compared settings. Table~\ref{tab:app_episode_ids} lists
the four-digit episode IDs within each task.

\begin{longtable}{@{}>{\raggedright\arraybackslash}p{0.21\linewidth}r>{\raggedright\arraybackslash}p{0.65\linewidth}@{}}
\caption{Reported ESI-Bench episode subset. IDs are listed within each subtask.}
\label{tab:app_episode_ids}\\
\toprule
Task & $n$ & Episode IDs \\
\midrule
\endfirsthead
\multicolumn{3}{l}{\small Table~\ref{tab:app_episode_ids} continued}\\
\toprule
Task & $n$ & Episode IDs \\
\midrule
\endhead
\bottomrule
\endfoot
Spatial Distance & 20 & 0781, 0787, 0788, 0822, 0835, 0838, 0841, 0844, 0851, 0856, 0857, 0866, 0867, 0869, 0871, 0874, 0877, 0894, 0907, 0910 \\[4pt]
Connectivity & 27 & 0078, 0081, 0086, 0087, 0089, 0090, 0092, 0096, 0098, 0100, 0101, 0102, 0103, 0107, 0108, 0109, 0115, 0120, 0122, 0123, 0125, 0126, 0128, 0131, 0132, 0133, 0134 \\[4pt]
Partial Occlusion & 20 & 1146, 1155, 1160, 1164, 1171, 1175, 1178, 1186, 1194, 1196, 1197, 1204, 1205, 1206, 1207, 1212, 1215, 1218, 1221, 1233 \\[4pt]
Action Order Inference & 20 & 0004, 0005, 0008, 0009, 0012, 0013, 0015, 0029, 0033, 0034, 0035, 0037, 0041, 0044, 0045, 0048, 0056, 0061, 0062, 0068 \\[4pt]
Physical Contact & 25 & 2804, 2810, 2812, 2819, 2824, 2829, 2833, 2835, 2837, 2841, 2850, 2854, 2855, 2862, 2868, 2873, 2877, 2881, 2890, 2895, 2896, 2898, 2899, 2916, 2921 \\[4pt]
Correspondence & 20 & 2923, 2937, 2941, 2948, 2952, 2955, 2958, 2959, 2964, 2967, 2968, 2974, 2978, 2981, 2982, 2985, 2988, 2989, 2990, 2998 \\[4pt]
Counting w/ Occlusion & 20 & 0398, 0402, 0403, 0404, 0405, 0407, 0408, 0410, 0411, 0412, 0413, 0414, 0415, 0417, 0418, 0419, 0420, 0421, 0425, 0427 \\[4pt]
Unobserved Change & 29 & 3357, 3362, 3366, 3372, 3379, 3382, 3384, 3386, 3391, 3395, 3401, 3402, 3407, 3413, 3414, 3419, 3427, 3437, 3442, 3444, 3448, 3457, 3460, 3464, 3471, 3480, 3499, 3500, 3501 \\[4pt]
Inclined Plane & 27 & 1677, 1679, 1680, 1683, 1686, 1689, 1690, 1695, 1697, 1698, 1701, 1702, 1708, 1710, 1712, 1713, 1714, 1715, 1717, 1721, 1725, 1726, 1727, 1730, 1733, 1734, 1737 \\[4pt]
\end{longtable}

\paragraph{Subset selection.}
All nine ESI-Bench tasks use sampling seed $42$. All our compared settings
use the same fixed episode lists in Table~\ref{tab:app_episode_ids}.

\subsection{EmbSpatial-Bench}
\label{app:embspatial}

We sample $300$ questions per relation with seed~42: Left, Right, Above,
Under, Far, and Close, giving $1{,}800$ questions. As in the main text,
\textsc{HiMeco} without evolution is compared with direct Gemini-3-Flash
API inference using the same questions and fixed visual inputs. Neither
setting uses active exploration or memory evolution. Equal category sizes
make macro and question-weighted accuracy equivalent before rounding.

\subsection{Metrics}
\label{app:metrics}

Answer accuracy uses the benchmark's task-dependent answer normalization and
scoring rule. It is distinct from learning success, which additionally
requires evidence eligibility. For task $j$ with $n_j$ episodes,
\begin{equation}
\mathrm{Acc}_j=\frac{1}{n_j}\sum_{i=1}^{n_j}
\mathbbm{1}[\hat y_{ji}=y_{ji}],\qquad
\mathrm{Acc}_{\mathrm{macro}}=\frac{1}{9}\sum_{j=1}^{9}\mathrm{Acc}_j.
\end{equation}
The main ESI-Bench table reports the unweighted mean of the nine task accuracies. Mean diagnostic scores cover all evaluated episodes. 
Primary failure-label rates in Figure~\ref{fig:failure}(a) are normalized by the total number of evaluated episodes. A correct trajectory may still receive a failure label when a planning anomaly is detected (Section~\ref{app:planning_score}).

Rounds~0 and~3 each use one candidate per episode. Intermediate-round
mean accuracy and best-of-$k$ accuracy are recorded separately. The sources
and sample counts of the comparison rows are described in
Section~\ref{app:settings} and Table~\ref{tab:main_results}.

\section{Program Execution and Evidence Checks}
\label{app:execution}

\subsection{Program representation and execution}
\label{app:ast}

The planner emits a complete JSON AST before execution. Primitive calls
assign outputs to named variables, and conditional statements branch on
tool outputs. Program structure is fixed before execution, while acquired
evidence determines the executed path.

Validation checks primitive names, references, output fields, selected
semantic types, statement count, branch depth, and an explicit
\texttt{return\_answer} on every path. One model repair attempt
is allowed, alongside deterministic normalization of evidence returns.
Programs are limited to $48$ statements and branch depth $4$.

Semantic primitives cover observation, navigation, visual and symbolic
reasoning, task interaction, and answer submission. Low-level camera
actions are executor internals; primitive calls can consume multiple
actions. Composite skills share the episode budget with the caller.
Captured physics frames are observations, not separate planner actions.
Skills return intermediate evidence; the outer program submits the answer.

\subsection{Evidence checks and answer consistency}
\label{app:contracts}

An \emph{evidence contract} is a set of rules that checks the format and
consistency of evidence returned by a visual primitive. A structured result can include an
answer, confidence, indices of the supporting views, visibility statements, per-view
explanations, and task-specific fields. A parsing or output-contract failure
can trigger one visual repair attempt. Valid JSON alone does not establish sufficient evidence.

The final answer is checked against the corresponding visual output using
its answer, evidence-view references, and contract fields. Programs return
the complete structured result to preserve this link.

For tasks that check the final evidence contract, a trajectory can serve
as a successful memory example only when
\begin{equation}
\ell(\tau)=\mathbbm{1}[\text{final evidence contract passes}],\qquad
s(\tau)=r_{\mathrm{task}}(\tau)\ell(\tau).
\end{equation}
This final gate applies to all evaluated tasks except Action Order Inference
and Physical Contact. Those two tasks retain visual-output checks and use
$\ell(\tau)=1$ when selecting successful memory examples.

\begin{longtable}{@{}>{\raggedright\arraybackslash}p{0.20\linewidth}>{\raggedright\arraybackslash}p{0.73\linewidth}@{}}
\caption{Task-specific visual evidence checks. Semantic fields are reported
by the visual model; checks assess their format and consistency.}
\label{tab:app_contracts}\\
\toprule
Task & Evidence checked by the visual module \\
\midrule
\endfirsthead
\multicolumn{2}{l}{\small Table~\ref{tab:app_contracts} continued}\\
\toprule Task & Evidence checked by the visual module \\
\midrule
\endhead
\bottomrule
\endfoot
Spatial Distance & A canonical candidate answer, visibility of the reference and both candidates, at least two cited views with per-view geometry, geometric viewpoint diversity, and no recorded contradictions. \\[3pt]
Connectivity & Identified regions and supporting cues; either evidence of a direct shared passage or explicit negative boundary/separation evidence. An intermediate region must not be mistaken for direct adjacency. \\[3pt]
Partial Occlusion & At least two cited views, tracking of the same target region, distinction between target and foreground blocker, an observed reduction in occlusion, identity cues from a revealing view, and geometric diversity. \\[3pt]
Action Order Inference & An ordered direction pair, a cited reference view, explicit selection of the correctly sized A/C candidates, and direction evidence consistent with the returned pair. \\[3pt]
Physical Contact & Evidence bound to the two named objects, a visible boundary or gap across views, rejection of ambiguous occlusion and misleading shadow/floor cues, and geometric diversity. \\[3pt]
Correspondence & A canonical answer among the allowed positions and structured evidence associating the reflected target with a real candidate while rejecting alternatives and contradictions. \\[3pt]
Counting w/ Occlusion & Structured object-instance evidence for the selected count, including identity association across views and occlusion-related evidence, rather than an unsupported scalar count. \\[3pt]
Unobserved Change & Evidence covering both temporal phases and nonempty box-specific evidence associated with the final answer. \\[3pt]
Inclined Plane & A canonical yes/no answer, nonempty evidence-view references, and motion evidence from frame observations; the record also carries stability, sliding, falling, and visibility fields. \\
\end{longtable}

For the geometric diversity check used by Spatial Distance, Partial
Occlusion, and Physical Contact, at least two distinct evidence positions
(rounded to three decimal places) are required
and either the maximum horizontal or vertical position separation must be
at least $0.08\,\mathrm{m}$. This is computed from solver-visible camera
poses. The check measures viewpoint separation.

\paragraph{Evidence repair and conflicting answers.}
The implementation can recover the final evidence record from an accepted
visual output. It also uses task-specific rules to resolve conflicting
answers. These operations occur before final benchmark scoring
and do not update persistent memory. They should be distinguished from
reflection between rounds. For Spatial Distance, conflicting accepted
estimates are compared using confidence, view/pose support, translation
baseline, elevation, and recorded contradictions. A conflicting estimate
replaces an already supported conclusion only if its evidence score exceeds
that conclusion by the configured margin $0.05$. These are heuristic
evidence checks; they do not query the ground-truth answer.

Contracts check evidence format, linkage, and internal consistency rather
than independently verifying scene geometry. Optional fields may be filled
from an accepted answer or frame description during normalization.

\section{Perception, Exploration, and Planning Diagnosis}
\label{app:diagnosis}

The diagnostic module returns three scores in $[0,1]$. Higher scores
indicate stronger evidence of failure. They need not sum to one and are
not calibrated probabilities. The rules below define each score.
Reflection uses all three scores, even when a figure shows only the
primary failure label.

\subsection{Exploration score}
\label{app:exploration_score}

Let $\mathcal P$ be the observed solver camera poses and $\mathcal G$ the
available reference camera positions. For each reference position $g_j$,
define its nearest observed position distance
$d_j=\min_{p\in\mathcal P}\|g_j-p\|_2$.
When reference distances are available, the position-coverage penalty is
\begin{equation}
c_{\mathrm{pose}}=
\frac{1}{|\mathcal G|}\sum_j
\operatorname{clip}\!\left(\frac{d_j-1.0}{1.5},0,1\right).
\end{equation}
Thus, distances at most $1\,\mathrm{m}$ incur no position penalty and
distances at least $2.5\,\mathrm{m}$ incur the maximum penalty. If no
reference distances can be computed, the fallback is $0$ when at least
three distinct positions were observed, and $0.7$ otherwise.

For diversity, positions and quaternion components are rounded to two
decimal places and distinct position--orientation pairs are counted as
$n_{\mathrm{pose}}$. Define
\begin{equation}
c_{\mathrm{div}}=\max(0,1-n_{\mathrm{pose}}/6),\qquad
c_{\mathrm{obs}}=\mathbbm{1}[|\mathcal P|\leq1\ \land\ n_{\mathrm{act}}>0].
\end{equation}
The exploration score is
\begin{equation}
u_{\mathrm{exp}}=
\min\!\left(1,0.6c_{\mathrm{pose}}+0.3c_{\mathrm{div}}+0.1c_{\mathrm{obs}}\right),
\label{eq:app_exploration}
\end{equation}
rounded to four decimal places. Reference-view coverage compares camera
positions only; pose diversity also counts camera orientations.

\subsection{Perception score}
\label{app:perception_score}

A correct benchmark answer receives $u_{\mathrm{per}}=0$. For an incorrect
answer, the verifier considers successfully executed visual calls, their
contract flags, any output from the symbolic \texttt{count\_unique} primitive, and the
exploration score. Table~\ref{tab:app_perception_rules} lists the rules in
evaluation order; the first applicable rule is used.

\begin{table}[htbp]
\centering
\small
\caption{Ordered rules for perception diagnosis after an incorrect answer.}
\label{tab:app_perception_rules}
\begin{tabular}{@{}>{\raggedright\arraybackslash}p{0.83\linewidth}r@{}}
\toprule
Condition & $u_{\mathrm{per}}$ \\
\midrule
A successful visual call reports an invalid output contract & 1.00 \\
Visual evidence is explicitly invalid and a symbolic count was returned & 0.95 \\
The last symbolic count disagrees with an available simulator-spawned target count & 1.00 \\
$u_{\mathrm{exp}}\leq0.35$, a visual call executed, and the answer is not unknown & 0.95 \\
$u_{\mathrm{exp}}\leq0.35$, a visual call executed, and the answer is unknown & 0.80 \\
A visual call executed, but the preceding rules do not apply & 0.55 \\
No successful visual call provides auditable evidence & 0.35 \\
\bottomrule
\end{tabular}
\end{table}

Unknown answers include empty outputs, \texttt{not sure}, \texttt{unknown},
\texttt{unsure}, and \texttt{none}. The count rule requires both simulator
target metadata and a symbolic count. Since correctness enters the score,
its mean summarizes internal diagnosis rather than an independent measure
of perception accuracy.

\subsection{Planning score and primary label}
\label{app:planning_score}

The rule-based planning score $u_{\mathrm{pla}}^{\mathrm{rule}}$ is determined
in the following order: program errors, a failed primitive, or incomplete
execution yield $1$; otherwise, a missing explicit return, terminal recovery,
or an invalid answer format yields $0.85$; otherwise, exceeding the action
budget yields $1$; an execution with none of these conditions receives $0$.

When enabled, a frozen language-model auditor examines the question, AST,
compact execution log, and branch log. It reviews primitive choice and
ordering, variable use, control flow, and supporting annotations. Its score
is clipped to $[0,1]$ and combined conservatively:
\begin{equation}
u_{\mathrm{pla}}=
\max\!\left(u_{\mathrm{pla}}^{\mathrm{rule}},u_{\mathrm{pla}}^{\mathrm{audit}}\right).
\end{equation}
The auditor cannot modify or continue the executed program. It uses
temperature $0$, top-$p=1$, and a configured output limit of $512$ tokens;
an unparsable audit falls back to the rule score.

The primary label is \texttt{none} when the answer is correct and the
planning score is zero. Otherwise, the largest rounded score determines
the label, with ties resolved in the order perception, exploration, planning.
All three scores remain available to the reflector.

\subsection{Candidate rewards and relative advantages}
\label{app:reward}

Candidate rewards use benchmark correctness, the mean of all three
diagnostic scores, and action efficiency:
\begin{equation}
R(\tau)=2r_{\mathrm{task}}(\tau)
-\frac{0.75}{3}\bigl(u_{\mathrm{per}}+u_{\mathrm{exp}}+u_{\mathrm{pla}}\bigr)
+0.1\left(1-\min\!\left(1,\frac{n_{\mathrm{act}}}{T_{\max}(q)}\right)\right).
\end{equation}
The reward uses the episode's task-specific budget $T_{\max}(q)$.
Subtracting the candidate-group mean gives $\mathrm{Adv}(\tau)$; it is
zero for a single-candidate group. Positive consolidation separately
requires $s(\tau)=1$, so reward ordering does not bypass evidence eligibility.

\section{Strategy and Skill Updates}
\label{app:memory}

\subsection{Strategy records and update operations}
\label{app:strategy}

Strategy records contain applicability conditions, lessons, recommended
program changes and primitive patterns, required actions, avoidance
constraints, and supporting candidate references. They also carry task,
diagnostic, creation-round, and optional episode identifiers. Updates
support addition, modification, deletion, and retention.

Consolidation combines three sources: the highest-advantage eligible
success per episode (ties favor fewer actions), a deterministic correction
when at least two candidates agree but none achieves learning success,
and updates proposed by the frozen reflector. Compact reflection inputs
include episode identifiers, candidate answers and outcomes, action
counts, advantages, all PEP scores, primary labels, and primitive
sequences; they omit images and the full simulator snapshot.
Reflection processes all selected episodes in batches of at most $20$.

Automatic capacity is $\max(20,2n_q)$ records per task, where $n_q$ is the
number of selected episode bundles. Compaction retains recent and supported
representatives by episode, experience type, and diagnostic dimension.
During adaptation, retrieval prefers matching episodes, task specificity,
recency, and support, with a diversity pass; at most five strategies are
provided. Final-round retrieval uses task-level entries.

\subsection{Creating and checking skills}
\label{app:skill_abstraction}

A composite skill stores its name, input types, program body, output,
conditions for use, source trajectories, and validation results.
To create a skill, the system renames local variables and replaces
eligible fixed arguments with \verb|$input.field| parameters. It preserves
variable dependencies and branches. Fixed strategies, directions, and
relations may remain in the skill body. Answer-return statements are removed.

The main extraction method groups programs that have the same body and
input types after parameterization. Each episode counts once toward a
pattern's support per update, even if several candidates contain it.
An episode counts as a success if at least one eligible occurrence succeeds.
By default, a pattern needs support from $m=2$ episodes and a success ratio
of at least $\theta_{\mathrm{syn}}=0.6$.

The system also extracts broader navigation and evidence patterns from
traces that pass evidence checks. It converts up to two executed
\texttt{explore\_scene} strategies into a standard form and adds a visual
primitive for the task. Connectivity has a separate extraction rule that
produces structured connectivity evidence. Repeated failures can also
produce trial routines that check object identities and change the
exploration strategy. Thus, skill extraction does not require an exact
match between program ASTs.

New skills start in a trial state: either shadow skills or experimental
corrective routines. Before trial execution, the system checks each skill's
name, primitives, variable references, structure, and task contract.
Each skill is limited to $12$ statements and branch depth $2$.
New skills become available in the next round. They still require validation
through actual calls, even when their source trajectories succeeded.

\subsection{Skill usage and status updates}
\label{app:lifecycle}

For each skill, the updater records which episodes called it, executed it
successfully, passed the evidence checks, and returned a correct answer
with passing evidence. Each episode counts at most once in each group per
update, even if the skill was called several times. Counts accumulate across
updates, so the same episode can count again in a later round. Accumulated
support therefore counts episode--round observations, not necessarily
distinct episodes.

Let the resulting counts be $N_{\mathrm{use}},N_{\mathrm{exec}},
N_{\mathrm{valid}},N_{\mathrm{succ}}$. Their conditional rates are
\begin{equation}
p_e=\frac{N_{\mathrm{exec}}}{N_{\mathrm{use}}},\qquad
p_c=\frac{N_{\mathrm{valid}}}{N_{\mathrm{exec}}},\qquad
p_s=\frac{N_{\mathrm{succ}}}{N_{\mathrm{valid}}},
\end{equation}
with zero used when a denominator is zero. Valid execution requires a
completed program, no execution error, and an answer in a valid format.

Skill status is updated using the rates and supporting episodes from the
\emph{current update}.
A skill is suspended if it fails to meet the task contract, its execution
success rate is below $0.9$, or its evidence pass rate is below the task threshold. If too few episodes pass the evidence checks in the current update, the
previous status is kept. With sufficient support, task success below $0.6$ also causes
suspension; otherwise the skill can become or remain active, subject to
configured comparison gates. A skill that was not called keeps its previous status.
Suspension changes eligibility without rewriting the validated body, and an
older trusted version may remain available.

\subsection{Skill quality and retrieval}
\label{app:skill_quality}

Quality statistics use
\begin{equation}
\rho(a,b)=\frac{a+\lambda\mu}{b+\lambda},\qquad
\mu=0.5,\quad\lambda=4.
\end{equation}
For successful calls that pass the evidence checks, let $v_i$ be the
recorded action ratio and $n$ the number of these recorded ratios.
The raw efficiency is
$\eta_{\mathrm{raw}}=\operatorname{clip}(1-\operatorname{median}_i v_i,0,1)$,
and the smoothed efficiency is
\begin{equation}
\tilde{\eta}=
\frac{n\eta_{\mathrm{raw}}+\lambda\mu}{n+\lambda},
\end{equation}
or $\mu$ when no successful invocation is available. The quality score is
\begin{equation}
Q(k)=w_s\rho(N_{\mathrm{succ}},N_{\mathrm{valid}})
+w_c\rho(N_{\mathrm{valid}},N_{\mathrm{exec}})
+w_e\rho(N_{\mathrm{exec}},N_{\mathrm{use}})
+w_a\tilde{\eta}.
\end{equation}
Skill efficiency uses an action-budget denominator of $30$ in the current
synthesis path; trajectory rewards use the task-specific budget.
Default weights are $(0.65,0.20,0.10,0.05)$, with task overrides listed in
Section~\ref{app:hyperparameters}.

The score $Q$ summarizes observed skill quality and supports candidate
selection and rollback. Retrieval first considers lifecycle status, then
empirical success and support. Final routing also checks execution
reliability, evidence validity, and task compatibility.

\subsection{Skill selection in the final round}
\label{app:skill_routing}

Final-round skill selection uses calls recorded before that round. It
checks the number of supporting episodes, execution and evidence pass
rates, task success, and whether the skill matches the task. A newly synthesized skill without execution evidence is
excluded. Eligible skills can guide evidence collection or be invoked when
applicable; a retrieved skill need not be called. These decisions use
stored outcomes, without access to the current episode's answer label.

\section{Connections to Geometric Perception and Continual Spatial Memory}
\label{app:perception_memory_context}

Reliable embodied spatial reasoning depends on both geometric
evidence acquisition and the ability to retain useful information
across observations. At the perception level, stereo matching
has explored slanted-plane cost aggregation \cite{wang2022spnet},
disparity-classification-aware context aggregation
\cite{wang2024dcanet}, and adaptive downsampling with disparity
alignment \cite{wang2025adstereo}. Robust geometric correspondence
has also benefited from self-supervised pseudo-labeling
\cite{wang2025dualnet} and the adaptation of visual foundation
models \cite{zhang2024formerstereo,wang2025smoestereo,wang2026smformer}.
At the memory level, \emph{Keep It in Mind}
\cite{wang2026keepitinmind} studies continual user-centric
spatial reasoning by maintaining structured memory from
streaming egocentric observations. These approaches address
complementary aspects of spatial intelligence: the stereo
methods focus on recovering geometry from image pairs, whereas
the memory-based approach supports object-location reasoning
over time. HiMeco targets a different challenge: rather than
introducing a stereo estimator or maintaining an explicit
object-location memory, it diagnoses perception, exploration,
and planning failures from embodied execution traces and
consolidates their lessons into reusable strategies and
executable skills.

\section{Implementation Settings}
\label{app:implementation}

\subsection{Model and program settings}
\label{app:hyperparameters}

The backbones are GPT-5.6-Terra and Gemini-3-Flash, accessed through APIs
with frozen parameters. Simulation runs on a server with eight NVIDIA
RTX A6000 GPUs. Table~\ref{tab:app_hyperparameters} lists reference settings;
task-specific values and decoding behavior are given below.

\begin{longtable}{@{}>{\raggedright\arraybackslash}p{0.55\linewidth}>{\raggedright\arraybackslash}p{0.38\linewidth}@{}}
\caption{Common reference-configuration settings. Task-dependent exceptions
are given below.}\label{tab:app_hyperparameters}\\
\toprule Setting & Value \\
\midrule
\endfirsthead
\multicolumn{2}{l}{\small Table~\ref{tab:app_hyperparameters} continued}\\
\toprule Setting & Value \\
\midrule
\endhead
\bottomrule
\endfoot
Round-0 planner temperature / top-$p$ & 0 / 1 \\
Intermediate planner temperature / top-$p$ & 0.7 / 0.9 \\
Final-generation temperature / top-$p$ & 0.7 / 0.9 in the reference runner \\
Planner output limit & 1,800 tokens \\
Program repair attempts after initial generation & 1 \\
Program statement limit / branch depth & 48 / 4 \\
Prompt demonstration limit & 15 \\
Visual output repair attempts & 1 \\
Planning auditor enabled & Yes \\
Planning auditor output limit & 512 tokens \\
Reflector output limit & 1,200 tokens \\
Reflector temperature / top-$p$ & 0 / 1 \\
Reflection episodes / batch size & All / 20 \\
Strategy retrieval limit & 5 \\
Skill retrieval limit & 5 \\
Automatic strategy capacity per task & $\max(20,2n_q)$ \\
Automatic skill capacity per task & $\max(10,n_q)$ \\
New skills per update / generic-skill limit & 5 / 3 \\
Skill statement limit / branch depth & 12 / 2 \\
Synthesis support / success threshold & 2 / 0.6 \\
Skill execution-reliability threshold & 0.9 \\
Quality prior mean / strength & 0.5 / 4 \\
\end{longtable}

\paragraph{Task-dependent settings.}
The task configurations set $T_{\max}=30$ for Spatial Distance,
Connectivity, Partial Occlusion, Action Order Inference, Physical Contact,
Correspondence, Counting w/ Occlusion, and Unobserved Change, and
$T_{\max}=12$ for Inclined Plane. The last task also uses four simulator
settling steps before its physical action in the current configuration.
Skill evidence validity is normally thresholded at $0.6$, with $0.7$ for
Spatial Distance. The current-update eligible-support threshold is normally
$2$, with $6$ for Spatial Distance.

The quality-weight override $(0.70,0.20,0.08,0.02)$ is present for Spatial
Distance, Action Order Inference, Physical Contact, Correspondence, and
Counting w/ Occlusion. Connectivity, Partial Occlusion, Unobserved Change,
and Inclined Plane use the default weights.

\paragraph{Decoding.}
Final-round generation uses temperature $0.7$ in the reference runner. Internal candidate seeds
depend on task, episode, round, and candidate index, while API requests
omit the seed. Provider-specific output limits can override the common
token limits in the table.

\subsection{Ablation settings}
\label{app:ablation_implementation}

All evolved variants in Table~\ref{tab:ablation_components} use the same episode
lists and four-round candidate schedule $(1,3,3,1)$; none uses same-episode
program replay in the reported evaluation. Each comparison changes the specified memory component
or reflection input. The first row reports the initial Round-0 baseline.

\paragraph{Memory ablations.}
The strategy-only variant disables skill use and updates; the skill-only
variant disables strategy use and updates.

\paragraph{Reflection-input ablation.}
The generic reflector reviews execution outcomes and suggests lessons and
program changes with PEP scores and failure-type labels removed from its
inputs. Reflection remains enabled. The intervention tests the contribution
of structured PEP feedback to reflection; it does not denote removal of
every diagnostic use elsewhere in the system. The $\mathcal U$ column in
Table~\ref{tab:ablation_components} denotes this structured reflection input.

\subsection{Execution records and timing}
\label{app:runtime}

Execution records include generated programs and distinguish retrieved
from invoked skills. Simulator retries are bounded, and
executions skipped after a crash are excluded from memory updates.

Trajectory time includes episode opening/reset, generation when needed,
execution, and diagnosis. It excludes round-level reflection and memory
updates, so it measures per-trajectory runtime rather than total adaptation
cost.

\FloatBarrier
\clearpage


\section{Prompt Templates}
\label{app:prompt_templates}
\label{app:prompts}

The following templates show how the planner, reflector, and planning auditor
are prompted. Instruction text is reproduced from the reference implementation;
line breaks and spacing are adjusted for readability. Colored placeholders
stand for runtime values. The planner payload is summarized below, and Spatial Distance illustrates
the task-specific instructions.

All boxes use the same headings: System instruction, Input fields, and
Instruction, with only the applicable parts shown.

\begin{prompttemplate}{Prompt 1: Program Planner}

\textnormal{\bfseries System instruction}\par\smallskip

You are the frozen Planner in a TANGO-style neuro-symbolic embodied agent.\par
Generate one COMPLETE executable JSON-AST program before execution begins.\par
Each executable line must call exactly one listed semantic primitive. Bind outputs with id and reference them as \$id or \$id.field.\par
Use if/then/else branches when later behavior depends on evidence. Low-level move\_\allowbreak{}*/turn\_\allowbreak{}* actions are executor internals and are forbidden in the program.\par
Respect every primitive output contract. Never compare whole structured outputs as numbers and never reference fields absent from output\_\allowbreak{}fields.\par
is\_\allowbreak{}found returns structured evidence (found, confidence, evidence\_\allowbreak{}valid), so it may be used as a branch condition or passed to is\_\allowbreak{}uncertain. Never replace that evidence object with a bare boolean before uncertainty testing.\par
For spatial distance and dimensional size, call one task-specific comparison primitive with all candidates and return its structured result.\par
If a retrieved learned\_\allowbreak{}composite skill matches the task and its input signature can be filled from the question, prefer invoking that skill instead of manually repeating its primitive body. If you choose not to invoke an available matching skill, the program metadata must include skill\_\allowbreak{}nonuse\_\allowbreak{}reason.\par
Treat retrieved\_\allowbreak{}strategy\_\allowbreak{}experiences as a wrong-problem notebook. Convert must\_\allowbreak{}do and recommended\_\allowbreak{}program\_\allowbreak{}pattern into executable primitive calls whenever they apply.\par
For any task affected by occlusion or viewpoint ambiguity, acquire geometrically different evidence before committing; an empty initial detection is uncertain evidence, not proof of absence.\par
Every control-flow path must execute return\_\allowbreak{}answer. Answers must be compatible with the listed task options.\par
Never use hidden simulator state, ground truth, oracle poses, or verifier data. Return one JSON object only.\par

\medskip

\end{prompttemplate}

The planner receives the task description and options, primitive signatures
and available skills, retrieved strategies, fixed demonstrations, candidate
index and strategy, and the initial and permitted reference images.
Requirements specify statement and action limits, answer-return coverage,
semantic type and evidence rules, memory-use policies, candidate diversity,
task-specific instructions, and an output-shape example.
An invalid program gets one repair attempt using the same prompt, the validation
error, and the instruction: \emph{Return a corrected complete program; do not
explain the error.}

\clearpage

\begin{prompttemplate}{Prompt 2: Strategy Reflection}

\textnormal{\bfseries System instruction}\par\smallskip

You update external strategy memory for a frozen neuro-symbolic Planner. Return JSON only.\par

\medskip

\textnormal{\bfseries Input fields}\par\smallskip

task\_\allowbreak{}key: \textcolor{promptvariable}{\textless{}TASK\_\allowbreak{}KEY\textgreater{}}\par
stage: \textcolor{promptvariable}{\textless{}ROUND\_\allowbreak{}INDEX\textgreater{}}\par
mode: \textcolor{promptvariable}{\textless{}REFLECTION\_\allowbreak{}MODE\textgreater{}}\par
candidates: \textcolor{promptvariable}{\textless{}CANDIDATE\_\allowbreak{}BUNDLES\textgreater{}}\par

\medskip

\textnormal{\bfseries Instruction}\par\smallskip

Compare same-episode programs and return JSON with experiences. Treat the experience memory as a wrong-problem notebook: write reusable semantic advantage about how to use existing primitives better, not episode-specific answers. Each experience must contain operation (add/modify/delete/keep), optional target\_\allowbreak{}id, failure\_\allowbreak{}dimension, when\_\allowbreak{}to\_apply, lesson, recommended\_\allowbreak{}program\_\allowbreak{}change, recommended\_\allowbreak{}program\_\allowbreak{}pattern, must\_\allowbreak{}do, avoid, and evidence\_\allowbreak{}candidate\_\allowbreak{}indices. If group\_\allowbreak{}pattern.all\_\allowbreak{}candidates\_\allowbreak{}learning\_\allowbreak{}failed (or legacy all\_\allowbreak{}candidates\_\allowbreak{}failed) and group\_\allowbreak{}pattern.identical\_\allowbreak{}final\_\allowbreak{}answer are true, create a systematic\_\allowbreak{}failure experience only when candidate\_\allowbreak{}count is at least two; a one-candidate baseline is not candidate collapse. Derive task-compatible recommendations from semantic\_\allowbreak{}primitive\_\allowbreak{}sequence and the observed failure bucket rather than assuming a hard-coded task template. Do not reveal ground-truth answers or episode-specific coordinates.\par

\medskip

\end{prompttemplate}

For the full method, candidate bundles contain outcomes, primitive sequences,
diagnostic scores, and relative advantages (Section~\ref{app:strategy}).
The generic-reflection ablation removes the PEP scores and labels as
specified in Section~\ref{app:ablation_implementation}.

\begin{prompttemplate}{Prompt 3: Planning Auditor}

\textnormal{\bfseries System instruction}\par\smallskip

You are an independent planning verifier. You cannot change or continue the Solver program. Return JSON only.\par

\medskip

\textnormal{\bfseries Input fields}\par\smallskip

question: \textcolor{promptvariable}{\textless{}QUESTION\textgreater{}}\par
program\_\allowbreak{}ast: \textcolor{promptvariable}{\textless{}EXECUTED\_\allowbreak{}PROGRAM\textgreater{}}\par
executed\_\allowbreak{}primitive\_\allowbreak{}logs: \textcolor{promptvariable}{\textless{}COMPACT\_\allowbreak{}PRIMITIVE\_\allowbreak{}LOGS\textgreater{}}\par
branch\_\allowbreak{}logs: \textcolor{promptvariable}{\textless{}BRANCH\_\allowbreak{}LOGS\textgreater{}}\par

\medskip

\textnormal{\bfseries Instruction}\par\smallskip

Audit primitive choice/order, variable use, branch logic, and whether the annotations support the plan. Return planning\_\allowbreak{}score in [0,1] where 1 means severe planning failure, plus issues and reasoning.\par

\medskip

\end{prompttemplate}

\clearpage

\begin{prompttemplate}{Prompt 4: Spatial Distance Policy}

\textnormal{\bfseries Instruction}\par\smallskip

Spatial Distance asks which candidate has the smaller physical ground-plane distance to the named reference object, not which candidate appears closer to the camera.\par
Perspective from one image is unreliable. Identify the reference and both candidates, then collect evidence from at least two genuinely different camera positions before committing.\par
Use lateral or depth parallax first; when the candidates remain at unfair image depths, move backward or sideways and compare again.\par
A modest real elevation change followed by downward pitch is an official task strategy for approximating a top-down layout view.\par
Call estimate\_\allowbreak{}distance once with both official answer options. Never compare two independent visual-assessment dictionaries or substitute apparent image size for object-to-object distance.\par
If visibility, geometric baseline, per-view evidence, or confidence is insufficient, branch to a geometrically different recovery view and call estimate\_\allowbreak{}distance again.\par
The final answer must be one exact official option; do not expose obj\_\allowbreak{}near/obj\_\allowbreak{}far metadata names in the program or reasoning.\par

\medskip

\end{prompttemplate}

This policy is supplied as part of the planner's task-specific instructions.
Other tasks use their own evidence and action rules. These fixed instructions
are present before memory evolution and are separate from learned strategies
and skills. The examples above cover program generation and retrospective
reflection; task-specific visual modules have separate prompts.


%% file: iclr2027_conference.bib
@article{hong2026esibench,
  title   = {{ESI-BENCH}: Towards Embodied Spatial Intelligence that Closes the Perception-Action Loop},
  author  = {Hong, Yining and Liu, Jiageng and Yin, Han and Li, Manling and Guibas, Leonidas and Fei-Fei, Li and Wu, Jiajun and Choi, Yejin},
  journal = {arXiv preprint arXiv:2605.18746
        
        },
  year    = {2026}
}

@inproceedings{du2024embspatial,
  title     = {{EmbSpatial-Bench}: Benchmarking Spatial Understanding for Embodied Tasks with Large Vision-Language Models},
  author    = {Du, Mengfei and Wu, Binhao and Li, Zejun and Huang, Xuanjing and Wei, Zhongyu},
  booktitle = {Proceedings of the 62nd Annual Meeting of the Association for Computational Linguistics (Volume 2: Short Papers)},
  pages     = {346--355},
  year      = {2024}
}

@inproceedings{ziliotto2025tango,
  title     = {{TANGO}: Training-free Embodied {AI} Agents for Open-world Tasks},
  author    = {Ziliotto, Filippo and Campari, Tommaso and Serafini, Luciano and Ballan, Lamberto},
  booktitle = {Proceedings of the IEEE/CVF Conference on Computer Vision and Pattern Recognition (CVPR)},
  year      = {2025}
}

@article{zhang2026sma,
  title   = {Spatial Memory Agent: Experience-Grounded Procedure Memory for Spatial Intelligence},
  author  = {Zhang, Haokai and Ding, Yuhang and Zhou, Yunshu and Du, Xinze and Zhang, Shengtao and Zhao, Zhiyue and Xi, Yuling and Chen, Hao},
  journal = {arXiv preprint arXiv:2608.12743
        
        },
  year    = {2026}
}

@article{tian2026nesy,
  title   = {Self-Evolving Neuro-Symbolic Skills for Tool-Augmented Spatial Reasoning},
  author  = {Tian, Shi-Yu and Wang, Zhuo-Xia and Zhu, Xuan-Yi and Zhou, Zhi and Yang, Xinwei and Yu, Kun-Yang and Yang, Ming and Chen, Yang and Li, Yu-Feng},
  journal = {arXiv preprint arXiv:2608.07955
        
        },
  year    = {2026}
}

@article{wang2023voyager,
  title   = {{VOYAGER}: An Open-Ended Embodied Agent with Large Language Models},
  author  = {Wang, Guanzhi and Xie, Yuqi and Jiang, Yunfan and Mandlekar, Ajay and Xiao, Chaowei and Zhu, Yuke and Fan, Linxi and Anandkumar, Anima},
  journal = {arXiv preprint arXiv:2305.16291
        
        },
  year    = {2023}
}

@article{wang2026shaper,
  title   = {Self-Evolving Embodied Agents via Skill-Harness Evolution},
  author  = {Wang, Peidong and Ma, Zhiming and Chang, Ying and Luo, Xufang and Yang, Xiaocui and Feng, Shi and Yang, Yuqing and Li, Dongsheng},
  journal = {arXiv preprint arXiv:2608.11350},
  year    = {2026}
}

@article{chen2026parallelworld,
  title={ParallelWorld: Test-Time Scaling for Embodied Reasoning},
  author={Chen, Min and Zhang, Shengjun and Li, Yuxin and Zhang, Zhang and Fei, Xin and Xia, Chong and Duan, Yueqi},
  journal={arXiv preprint arXiv:2608.22971},
  year={2026}
}

@article{ju2026embodiskill,
  title   = {{EmbodiSkill}: Skill-Aware Reflection for Self-Evolving Embodied Agents},
  author  = {Ju, Ruofei and Wang, Xinrui and Ding, Xin and Yang, Yifan and Wu, Hao and Jiang, Shiqi and Zhang, Qianxi and Wen, Hao and Li, Xiangyu and Wang, Weijun and Li, Kun and Liu, Yunxin and Dai, Haipeng and Wang, Wei and Cao, Ting},
  journal = {arXiv preprint arXiv:2605.10332},
  year    = {2026}
}

@article{wang2026metaskill,
  title   = {{MetaSkill-Evolve}: Recursive Self-Improvement of {LLM} Agents via Two-Timescale Meta-Skill Evolution},
  author  = {Wang, Zefeng and Yan, Minxi and Bi, Jinhe and Yan, Sikuan and Tresp, Volker and Ma, Yunpu},
  journal = {arXiv preprint arXiv:2607.05297},
  year    = {2026}
}

@inproceedings{zhang2026dgm,
  title     = {Darwin {G{\"o}del} Machine: Open-Ended Evolution of Self-Improving Agents},
  author    = {Zhang, Jenny and Hu, Shengran and Lu, Cong and Lange, Robert and Clune, Jeff},
  booktitle = {International Conference on Learning Representations (ICLR)},
  year      = {2026}
}

@article{youtu2025tfgrpo,
  title   = {Training-Free Group Relative Policy Optimization},
  author  = {{Youtu-Agent Team}},
  journal = {arXiv preprint arXiv:2510.08191},
  year    = {2025}
}

@inproceedings{shu2026remem,
  title     = {{REMem}: Reasoning with Episodic Memory in Language Agent},
  author    = {Yiheng Shu and Padmaja Jonnalagedda and Xiang Gao and Bernal Jimenez Gutierrez and Weijian Qi and Kamalika Das and Huan Sun and Yu Su},
  booktitle = {International Conference on Learning Representations},
  year      = {2026}
}

@inproceedings{zhu2025llava3d,
  title     = {{LLaVA-3D}: A Simple yet Effective Pathway to Empowering {LMMs} with {3D} Capabilities},
  author    = {Chenming Zhu and Tai Wang and Wenwei Zhang and Jiangmiao Pang and Xihui Liu},
  booktitle = {Proceedings of the IEEE/CVF International Conference on Computer Vision},
  year      = {2025}
}

@inproceedings{xu2026multigranularity,
  title     = {From Single to Multi-Granularity: Toward Long-Term Memory Association and Selection of Conversational Agents},
  author    = {Derong Xu and Yi Wen and Pengyue Jia and Yingyi Zhang and Wenlin Zhang and Yichao Wang and Huifeng Guo and Ruiming Tang and Xiangyu Zhao and Enhong Chen and Tong Xu},
  booktitle = {International Conference on Learning Representations},
  year      = {2026}
}

@inproceedings{zeng2026janusvln,
  title     = {{JanusVLN}: Decoupling Semantics and Spatiality with Dual Implicit Memory for Vision-Language Navigation},
  author    = {Shuang Zeng and Dekang Qi and Xinyuan Chang and Feng Xiong and Shichao Xie and Xiaolong Wu and Shiyi Liang and Mu Xu and Xing Wei},
  booktitle = {International Conference on Learning Representations},
  year      = {2026}
}

@inproceedings{qi2026armor,
  title     = {Self-Refining Vision Language Model for Robotic Failure Detection and Reasoning},
  author    = {Carl Qi and Xiaojie Wang and Silong Yong and Stephen Sheng and Huitan Mao and Sriram Srinivasan and Mani Nambi and Amy Zhang and Yesh Dattatreya},
  booktitle = {International Conference on Learning Representations},
  year      = {2026}
}

@inproceedings{gong2026space10,
  title     = {{SpaCE-10}: A Comprehensive Benchmark for Multimodal Large Language Models in Compositional Spatial Intelligence},
  author    = {Ziyang Gong and Wenhao Li and Xianzheng Ma and Songyuan Li and Zhaokai Wang and Songze Li and Jiayi Ji and Xue Yang and Gen Luo and Junchi Yan and Rongrong Ji},
  booktitle = {International Conference on Learning Representations},
  year      = {2026}
}

@inproceedings{feng2026seeing,
  title     = {Seeing Across Views: Benchmarking Spatial Reasoning of Vision-Language Models in Robotic Scenes},
  author    = {Zhiyuan Feng and Zhaolu Kang and Qijie Wang and Zhiying Du and Jiongrui Yan and Shubin Shi and Chengbo Yuan and Huizhi Liang and Yu Deng and Qixiu Li and Rushuai Yang and Ruichuan An and Leqi Zheng and Weijie Wang and Shuang Chen and Sicheng Xu and Yaobo Liang and Jiaolong Yang and Baining Guo},
  booktitle = {International Conference on Learning Representations},
  year      = {2026}
}

@inproceedings{xu2025amem,
  title     = {{A-Mem}: Agentic Memory for {LLM} Agents},
  author    = {Wujiang Xu and Zujie Liang and Kai Mei and Hang Gao and Juntao Tan and Yongfeng Zhang},
  booktitle = {Advances in Neural Information Processing Systems},
  volume    = {38},
  year      = {2025}
}

@inproceedings{wang2025awm,
  title     = {Agent Workflow Memory},
  author    = {Zora Zhiruo Wang and Jiayuan Mao and Daniel Fried and Graham Neubig},
  booktitle = {Proceedings of the 42nd International Conference on Machine Learning},
  series    = {Proceedings of Machine Learning Research},
  volume    = {267},
  pages     = {63897--63911},
  publisher = {PMLR},
  year      = {2025}
}

@inproceedings{liu2025cer,
  title     = {Contextual Experience Replay for Self-Improvement of Language Agents},
  author    = {Yitao Liu and Chenglei Si and Karthik R. Narasimhan and Shunyu Yao},
  booktitle = {Proceedings of the 63rd Annual Meeting of the Association for Computational Linguistics (Volume 1: Long Papers)},
  pages     = {14179--14198},
  publisher = {Association for Computational Linguistics},
  year      = {2025}
}

@inproceedings{duan2025aha,
  title     = {{AHA}: A Vision-Language-Model for Detecting and Reasoning Over Failures in Robotic Manipulation},
  author    = {Jiafei Duan and Wilbert Pumacay and Nishanth Kumar and Yi Ru Wang and Shulin Tian and Wentao Yuan and Ranjay Krishna and Dieter Fox and Ajay Mandlekar and Yijie Guo},
  booktitle = {International Conference on Learning Representations},
  year      = {2025}
}

@inproceedings{yang2025thinking,
  title     = {Thinking in Space: How Multimodal Large Language Models See, Remember, and Recall Spaces},
  author    = {Jihan Yang and Shusheng Yang and Anjali W. Gupta and Rilyn Han and Li Fei-Fei and Saining Xie},
  booktitle = {Proceedings of the IEEE/CVF Conference on Computer Vision and Pattern Recognition},
  year      = {2025}
}

@inproceedings{song2025robospatial,
  title     = {{RoboSpatial}: Teaching Spatial Understanding to {2D} and {3D} Vision-Language Models for Robotics},
  author    = {Chan Hee Song and Valts Blukis and Jonathan Tremblay and Stephen Tyree and Yu Su and Stan Birchfield},
  booktitle = {Proceedings of the IEEE/CVF Conference on Computer Vision and Pattern Recognition},
  year      = {2025}
}

@inproceedings{wang2025metavqa,
  title     = {Embodied Scene Understanding for Vision Language Models via {MetaVQA}},
  author    = {Weizhen Wang and Chenda Duan and Zhenghao Peng and Yuxin Liu and Bolei Zhou},
  booktitle = {Proceedings of the IEEE/CVF Conference on Computer Vision and Pattern Recognition},
  year      = {2025}
}

@inproceedings{kim2025flare,
  title     = {Multi-Modal Grounded Planning and Efficient Replanning for Learning Embodied Agents with a Few Examples},
  author    = {Taewoong Kim and Byeonghwi Kim and Jonghyun Choi},
  booktitle = {Proceedings of the AAAI Conference on Artificial Intelligence},
  volume    = {39},
  pages     = {4329--4337},
  year      = {2025}
}

@inproceedings{zhang2026memgen,
  title     = {{MemGen}: Weaving Generative Latent Memory for Self-Evolving Agents},
  author    = {Guibin Zhang and Muxin Fu and Shuicheng Yan},
  booktitle = {International Conference on Learning Representations},
  year      = {2026}
}

@article{zheng2025esceme,
  title   = {{ESceme}: Vision-and-Language Navigation with Episodic Scene Memory},
  author  = {Qi Zheng and Daqing Liu and Chaoyue Wang and Jing Zhang and Dadong Wang and Dacheng Tao},
  journal = {International Journal of Computer Vision},
  volume  = {133},
  pages   = {254--274},
  year    = {2025},
  doi     = {10.1007/s11263-024-02159-8}
}

@inproceedings{li2024eai,
  title     = {Embodied Agent Interface: Benchmarking {LLMs} for Embodied Decision Making},
  author    = {Manling Li and Shiyu Zhao and Qineng Wang and Kangrui Wang and Yu Zhou and Sanjana Srivastava and Cem Gokmen and Tony Lee and Li Erran Li and Ruohan Zhang and Weiyu Liu and Percy Liang and Li Fei-Fei and Jiayuan Mao and Jiajun Wu},
  booktitle = {Advances in Neural Information Processing Systems},
  volume    = {37},
  year      = {2024}
}

@inproceedings{cheng2024spatialrgpt,
  title     = {{SpatialRGPT}: Grounded Spatial Reasoning in Vision-Language Models},
  author    = {An-Chieh Cheng and Hongxu Yin and Yang Fu and Qiushan Guo and Ruihan Yang and Jan Kautz and Xiaolong Wang and Sifei Liu},
  booktitle = {Advances in Neural Information Processing Systems},
  volume    = {37},
  year      = {2024}
}

@inproceedings{majumdar2024openeqa,
  title     = {{OpenEQA}: Embodied Question Answering in the Era of Foundation Models},
  author    = {Arjun Majumdar and Anurag Ajay and Xiaohan Zhang and Pranav Putta and Sriram Yenamandra and Mikael Henaff and Sneha Silwal and Paul Mcvay and Oleksandr Maksymets and Sergio Arnaud and Karmesh Yadav and Qiyang Li and Ben Newman and Mohit Sharma and Vincent Berges and Shiqi Zhang and Pulkit Agrawal and Yonatan Bisk and Dhruv Batra and Mrinal Kalakrishnan and Franziska Meier and Chris Paxton and Alexander Sax and Aravind Rajeswaran},
  booktitle = {Proceedings of the IEEE/CVF Conference on Computer Vision and Pattern Recognition},
  pages     = {16488--16498},
  year      = {2024}
}

@article{liao2024learning,
  title   = {Learning, Fast and Slow: Single- and Many-Shot Learning in the Hippocampus},
  author  = {Liao, Zhenrui and Losonczy, Attila},
  journal = {Annual Review of Neuroscience},
  volume  = {47},
  pages   = {187--209},
  year    = {2024}
}

@article{huang2024replay,
  title   = {Replay-Triggered Brain-Wide Activation in Humans},
  author  = {Huang, Qi and Xiao, Zhibing and Yu, Qianqian and Luo, Yuejia and Xu, Jiahua and Qu, Yukun and Dolan, Raymond J. and Behrens, Timothy E. J. and Liu, Yunzhe},
  journal = {Nature Communications},
  volume  = {15},
  pages   = {7185},
  year    = {2024}
}

@article{jensen2024recurrent,
  title   = {A Recurrent Network Model of Planning Explains Hippocampal Replay and Human Behavior},
  author  = {Jensen, Kristopher T. and Hennequin, Guillaume and Mattar, Marcelo G.},
  journal = {Nature Neuroscience},
  volume  = {27},
  number  = {7},
  pages   = {1340--1348},
  year    = {2024}
}

@article{fleming2026integrative,
  title   = {Towards an Integrative Neuroscience of Metacognition},
  author  = {Fleming, Stephen M.},
  journal = {Nature Reviews Neuroscience},
  year    = {2026}
}

@inproceedings{shi2026memoryvla,
  title     = {{MemoryVLA}: Perceptual-Cognitive Memory in Vision-Language-Action Models for Robotic Manipulation},
  author    = {Hao Shi and Bin Xie and Yingfei Liu and Lin Sun and Fengrong Liu and Tiancai Wang and Erjin Zhou and Haoqiang Fan and Xiangyu Zhang and Gao Huang},
  booktitle = {International Conference on Learning Representations},
  year      = {2026}
}

@article{wang2022spnet,
  title   = {{SPNet}: Learning Stereo Matching With Slanted Plane Aggregation},
  author  = {Wang, Yun and Wang, Longguang and Wang, Hanyun and Guo, Yulan},
  journal = {IEEE Robotics and Automation Letters},
  volume  = {7},
  number  = {3},
  pages   = {6258--6265},
  year    = {2022},
  doi     = {10.1109/LRA.2022.3164755}
}

@article{wang2024dcanet,
  title   = {Cost Volume Aggregation in Stereo Matching Revisited: A Disparity Classification Perspective},
  author  = {Wang, Yun and Wang, Longguang and Li, Kunhong and Zhang, Yongjian and Wu, Dapeng Oliver and Guo, Yulan},
  journal = {IEEE Transactions on Image Processing},
  volume  = {33},
  pages   = {6425--6438},
  year    = {2024},
  doi     = {10.1109/TIP.2024.3484251}
}

@article{wang2025adstereo,
  title   = {{ADStereo}: Efficient Stereo Matching With Adaptive Downsampling and Disparity Alignment},
  author  = {Wang, Yun and Li, Kunhong and Wang, Longguang and Hu, Junjie and Wu, Dapeng Oliver and Guo, Yulan},
  journal = {IEEE Transactions on Image Processing},
  volume  = {34},
  pages   = {1204--1218},
  year    = {2025},
  doi     = {10.1109/TIP.2025.3540282}
}

@inproceedings{wang2025dualnet,
  title     = {{DualNet}: Robust Self-Supervised Stereo Matching with Pseudo-Label Supervision},
  author    = {Wang, Yun and Zheng, Jiahao and Zhang, Chenghao and Zhang, Zhanjie and Li, Kunhong and Zhang, Yongjian and Hu, Junjie},
  booktitle = {Proceedings of the AAAI Conference on Artificial Intelligence},
  volume    = {39},
  number    = {8},
  pages     = {8178--8186},
  year      = {2025},
  doi       = {10.1609/aaai.v39i8.32882}
}

@inproceedings{zhang2024formerstereo,
  title     = {Learning Representations from Foundation Models for Domain Generalized Stereo Matching},
  author    = {Zhang, Yongjian and Wang, Longguang and Li, Kunhong and Wang, Yun and Guo, Yulan},
  booktitle = {European Conference on Computer Vision},
  year      = {2024},
  doi       = {10.1007/978-3-031-72946-1_9}
}

@inproceedings{wang2025smoestereo,
  title     = {Learning Robust Stereo Matching in the Wild with Selective Mixture-of-Experts},
  author    = {Wang, Yun and Wang, Longguang and Zhang, Chenghao and Zhang, Yongjian and Zhang, Zhanjie and Ma, Ao and Fan, Chenyou and Lam, Tin Lun and Hu, Junjie},
  booktitle = {Proceedings of the IEEE/CVF International Conference on Computer Vision},
  pages     = {21276--21287},
  year      = {2025},
  doi       = {10.1109/ICCV51701.2025.01976}
}

@article{wang2026smformer,
  title   = {{SMFormer}: Empowering Self-Supervised Stereo Matching via Foundation Models and Data Augmentation},
  author  = {Wang, Yun and Yang, Zhengjie and Zheng, Jiahao and Zhang, Zhanjie and Wu, Dapeng Oliver and Guo, Yulan},
  journal = {IEEE Transactions on Image Processing},
  volume  = {35},
  pages   = {4494--4508},
  year    = {2026},
  doi     = {10.1109/TIP.2026.3684415}
}

@inproceedings{wang2026keepitinmind,
  title     = {{Keep It in Mind}: User Centric Continual Spatial Intelligence Reasoning in Egocentric Video Streams},
  author    = {Wang, Yun and Xiao, Junbin and Lyu, Han and Wang, Yifan and Zuo, Jing and Zhang, Zhanjie and Huang, Hong and Wu, Dapeng and Yao, Angela},
  booktitle = {Proceedings of the 43rd International Conference on Machine Learning},
  series    = {Proceedings of Machine Learning Research},
  volume    = {306},
  pages     = {131373--131417},
  year      = {2026},
  publisher = {PMLR},
  url       = {https://proceedings.mlr.press/v306/wang26je.html}
}
